\documentclass[10pt,twocolumn,letterpaper]{article}

\usepackage[pagenumbers]{cvpr} 

\usepackage{siunitx}
\usepackage{multirow}
\usepackage{pifont}

\newcommand{\minisection}[1]{\vspace{0.025in}\noindent{\bf #1}\hspace{0.10em}}
\newcommand{\red}[1]{{\color{red}#1}}

\newcommand{\tit}[1]{\smallbreak\noindent\textbf{#1 }}

\newcommand{\tinytit}[1]{\noindent\textbf{#1}}

\definecolor{ao(english)}{rgb}{0.0, 0.5, 0.0}

\newcommand{\dgreen}[1]{\textcolor{ao(english)}{#1}}

\newcommand{\cmark}{\ding{51}}%
\newcommand{\greencmark}{\dgreen{\cmark}}
\newcommand{\xmark}{\ding{55}}%
\newcommand{\redxmark}{\red{\xmark}}

\definecolor{cvprblue}{rgb}{0.21,0.49,0.74}
\usepackage[pagebackref,breaklinks,colorlinks,allcolors=cvprblue]{hyperref}

\def\confName{CVPR}
\def\confYear{2027}

\title{Dual-Tuned Experts: Classifier-First Training for Model Merging}

\author{
  Christos Georgakilas$^{1,2}$\thanks{Corresponding author: \texttt{cgeorgakilas@cvc.uab.cat}} \quad
  Aniello Panariello$^{3}$ \quad
  Samir El Karrat Moreno$^{2}$ \\
  Simone Calderara$^{3}$ \quad
  Dimosthenis Karatzas$^{1,2}$ \quad
  Joost van de Weijer$^{1,2}$ \\[0.8em]
  $^{1}$ Computer Vision Center, Barcelona, Spain \\
  $^{2}$ Universitat Autònoma de Barcelona, Barcelona, Spain \\
  $^{3}$ AImageLab, University of Modena and Reggio Emilia, Italy
}

\begin{document}
\maketitle
\begin{abstract}
Model merging combines task-specific experts derived from a shared foundation model into a model that serves multiple tasks. While much research focuses on the merging operator, the training procedure used to obtain the experts is another important choice. We study classifier-first expert training under the standard task-aware merging evaluation protocol. 
We first show that classifier-only adaptation is a strong baseline for task-aware model merging, becoming particularly competitive as the number of tasks increases.
However, for some tasks, it can leave notable performance gaps relative to backbone fine-tuned experts. Motivated by this observation, we introduce Dual-Tuned Experts (DTEs), a two-stage procedure that first learns a classifier on the frozen pre-trained backbone and then adapts the backbone while keeping that classifier fixed. 
The resulting experts can be combined with existing merging operators to improve merged-system accuracy. Our analyses of mode-connectivity and representation-preservation support the merged performance benefits obtained with DTEs.
The same formulation applies to both vision and language model merging, accommodating learned text prompts in CLIP and linear classification heads in DINOv2 and Llama~3 8B.
Across full-parameter vision fine-tuning and LoRA-based vision and language adaptation, DTEs consistently improve merged-system accuracy with existing merging operators. 
\end{abstract}    
\section{Introduction}
\label{sec:intro}

In recent years, large-scale foundation models have been widely adopted in computer vision, enabling adaptation to diverse downstream tasks through fine-tuning. The resulting collection of experts motivates model merging~\cite{ilharco2023editing}: combining checkpoints derived from a shared foundation model without joint retraining. Model merging has been studied across computer vision, large language models, and medical imaging~\cite{yang2026model,lumetti2025u}.

Early approaches to model merging focused on combining models trained on the same dataset with different hyperparameter configurations~\cite{wortsman2022model}. Ilharco \etal~\cite{ilharco2023editing} introduced task vectors, defined as the difference between a fine-tuned model and its base model, and showed that they can be combined through arithmetic in parameter space. Subsequent work reduces interference at merge time through sign resolution~\cite{yadav2023tiesmerging}, parameter selection~\cite{MarczakTTC24}, task-localization masks~\cite{wang2024consensus}, adaptive weighting~\cite{yang2024adamerging}, and structured decompositions of task vectors~\cite{tsv,marczak2025task,rinaldi2026transporting}.

Expert training is also an active design axis. Tangent-space fine-tuning~\cite{ortizjimenez2023tangent}, partial linearization for parameter-efficient fusion~\cite{tang2024partial}, sparse task-localized fine-tuning~\cite{iurada2025talos}, sharpness-aware fine-tuning~\cite{lee2025saft}, and merge-aware optimization~\cite{yang2026mergopt} all shape the adapted expert or its update to improve later composition. Our contribution is complementary: given a classification model, we ask how adapting the classifier first changes the backbone update learned in a subsequent, otherwise conventional fine-tuning stage and whether this benefits downstream merging performance.

 Classifier-only adaptation offers a simple alternative to backbone fine-tuning under task-aware evaluation. A task-specific classifier is trained on the frozen pre-trained representations and selected using the known task identity. For CLIP, this can be implemented through prompt learning; for vision-only backbones and language models, it can be implemented through a learnable linear head. This strategy avoids backbone weight merging and provides an important comparison for merged systems. However, classifier-only adaptation may be insufficient for tasks that benefit from changes to the underlying representations.

Motivated by these observations, we introduce Dual-Tuned Experts (DTEs), a two-stage expert-training recipe that separates classifier training from backbone fine-tuning. In the first stage, only the task classifier is adapted on frozen backbone features. In the second stage, the trained classifier is frozen and only the backbone is fine-tuned, yielding a task-specific pair of backbone weights and classifier. Our intuition is that a classifier already aligned with the pre-trained features may reduce the representational changes needed during backbone adaptation. Since model merging combines the resulting backbone updates, these changes could influence downstream merging performance.

Head-first adaptation is well established in transfer learning, with early works demonstrating the effectiveness of frozen features for downstream tasks~\cite{donahue2014decaf,oquab2014learning}. More recently, LP-FT~\cite{kumar2022lpft} initializes the head with a linear probe before full fine-tuning to reduce feature distortion and improve out-of-distribution robustness, and FLYP~\cite{goyal2023flyp} shows that the fine-tuning objective matters for CLIP. 

These works evaluate each adapted model in isolation. We are the first to systematically investigate classifier-first training in the context of model merging, studying how this affects the combination of independently trained backbones into a single model.
We evaluate DTEs with existing merging methods, including Weight Averaging~\cite{wortsman2022model}, Task Arithmetic~\cite{ilharco2023editing}, TIES~\cite{yadav2023tiesmerging}, TSV-M~\cite{tsv} and Iso-C~\cite{marczak2025task}.

The main contributions of this paper are:
\begin{itemize}
    \item We establish classifier-only adaptation as an important baseline for task-aware model merging, showing that prompt learning is particularly competitive on many-task CLIP benchmarks.

    \item We study Dual-Tuned Experts (DTEs), a unified classifier-first training procedure that learns a task-specific classifier before adapting the backbone with that classifier fixed. We analyze representation preservation and interpolation behavior.

    \item We evaluate DTEs on CLIP, DINOv2, and Llama~3 8B using existing merging operators. Across the reported comparisons, DTE systems improve merged-model accuracy under their retained task-specific classifiers. Notably, on DINOv2 with 20 tasks, DTE improves TSV-M accuracy by $25.17$ percentage points while standalone expert accuracy only marginally improves, demonstrating substantial gains in merged-systems performance.
\end{itemize} 
\section{Related Work}
\label{sec:rel}

\tit{Merging operators.}
Model merging combines fine-tuned checkpoints from a shared pre-trained model without joint retraining~\cite{matena2021merging,wortsman2022model}. Model Soups establishes weight averaging~\cite{wortsman2022model}, and Task Arithmetic represents experts as composable task vectors~\cite{ilharco2023editing}. Extensive literature then modifies the merge itself: TIES resolves sign conflicts~\cite{yadav2023tiesmerging}; Consensus Merging identifies harmful task-vector components~\cite{wang2024consensus}; AdaMerging learns task-dependent merge weights~\cite{yang2024adamerging}; EMR-MERGING elects a unified model and adds lightweight task-specific masks and rescalers without tuning~\cite{huang2024emr}; and TSV-M, Iso-C, Core Space, and KnOTS exploit update structure or low-rank subspaces~\cite{tsv,marczak2025task,panariello2025accurate,stoica2024knots}. DTE is orthogonal to these operators: it alters how experts are adapted and can be paired with a chosen merging rule.

\tit{Training experts for composition.}
Several works explicitly modify expert training or parameterization to improve downstream editing or merging. Tangent-space fine-tuning amplifies weight disentanglement through linearization~\cite{ortizjimenez2023tangent}; partial linearization applies this idea to adapters~\cite{tang2024partial}; TaLoS promotes localized sparse updates~\cite{iurada2025talos}; and sharpness-aware fine-tuning seeks both accurate experts and lower interference~\cite{lee2025saft}. MergOPT directly optimizes experts for stability under merge-induced perturbations~\cite{yang2026mergopt}. DTE uses downstream classification supervision to adapt the classifier before updating the backbone, without access to other experts or merge-induced perturbations. The classifier is implemented through learned prompts in CLIP and linear heads in DINOv2 and Llama~3. DTEs can be combined with existing merging operators.

\tit{Prompt and head-first adaptation.}
Prompt learning adapts frozen pre-trained models by optimizing a small set of learnable input tokens~\cite{jia2022visual,liu2022p}. In CLIP, CoOp replaces handcrafted templates with learnable context tokens~\cite{zhou2022learning}; CoCoOp, MaPLe, and PromptSRC respectively improve conditional generalization, multimodal prompting, and prompt regularization~\cite{zhou2022conditional,khattak2023maple,khattak2023self}. DTE is not a new prompt-learning method. Conceptually, it is more related to LP-FT, which learns a classifier before full fine-tuning to reduce feature distortion~\cite{kumar2022lpft}, and to FLYP, which retains CLIP's contrastive image--text adaptation objective~\cite{goyal2023flyp}. However, we focus on the performance of independently adapted experts after merging, including the consequences of keeping the learned classifier fixed during backbone adaptation.
\section{Preliminaries}
\label{sec:prelim}

\subsection{Model Merging}
Given a pre-trained backbone with parameters $\theta_0$, we independently adapt it to $T$ tasks, obtaining task-specific parameters $\{\theta_1,\ldots,\theta_T\}$. Model merging combines these experts without joint training or access to the original training data. Task Arithmetic~\cite{ilharco2023editing} defines task vectors as $\Delta\theta_i=\theta_i-\theta_0$ and computes the merged parameters as:
\begin{equation}
    \theta_{\text{merged}} = \theta_0 +
    \lambda \sum_{i=1}^{T}\Delta\theta_i,
\end{equation}
where $\lambda$ is selected on a held-out validation set.

The effectiveness of weight-space merging is related to flat loss regions~\cite{mehta2023empirical} and mode connectivity~\cite{frankle2020linear,mirzadeh2020linear}. However, independently learned updates may interfere when combined, motivating approaches that improve their compatibility.

\begin{figure*}[t]
    \centering
    \includegraphics[width=\textwidth]{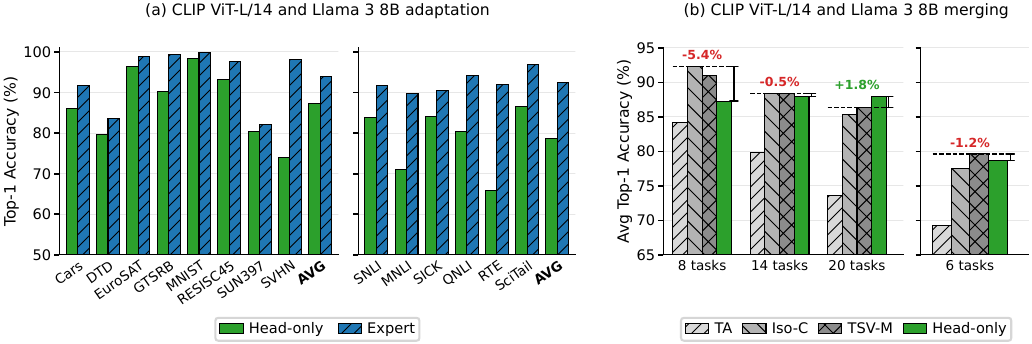}
    \caption{\textbf{(a)} Performance of CLIP ViT-L/14 on the 8-vision tasks and Llama~3 8B on 6 NLI tasks. Results include classifier-only and full model (Llama) or backbone (CLIP) training. \textbf{(b)} Average performance of merging CLIP ViT-L/14 across 8, 14, and 20 vision tasks and Llama~3 8B across 6 NLI tasks. Results compare Task Arithmetic (TA), Iso-C, and TSV-M against a classifier-only training baseline.}
    \label{fig:fig1long}
\end{figure*}
\subsection{Backbone-Classifier Framework}
\label{sec:backbone_classifier}

We consider classification models composed of a \emph{backbone}, which extracts input representations, and a \emph{classifier}, which maps these representations to task-specific predictions. Formally, given an input $x$, the model computes:
\begin{equation}
    \mathbf{h} = f_\theta(x), \qquad
    \mathbf{z} = g_\phi(\mathbf{h}),
\end{equation}
where $f_\theta$ denotes the backbone parameterized by $\theta$, $g_\phi$ denotes the classifier parameterized by $\phi$, and $\mathbf{z}\in\mathbb{R}^{C}$ contains the classification scores for $C$ classes.

For a collection of $T$ tasks, each task $i$ is associated with a classifier $g_{\phi_i}$ and an adapted backbone $f_{\theta_i}$. In classification-based model merging, only the backbone parameters are merged, while task-specific classifiers are retained. Given the known task identity $i$, inference uses:
\begin{equation}
    \mathbf{z}_i =
    g_{\phi_i}\!\left(f_{\theta_{\text{merged}}}(x)\right).
    \label{eq:merged_classifier}
\end{equation}

We consider three instantiations: CLIP, which uses text-derived class prototypes; DINOv2, which employs a linear classifier over visual features; and decoder-only LLMs, which classify contextualized language representations.

\subsection{CLIP for Classification}
CLIP~\cite{radford2021learning} consists of a vision encoder $f_v$ and a text encoder $f_t$, mapping images and text into a shared embedding space. The vision encoder acts as the backbone, while the text encoder generates class prototypes from prompts $\{p_c\}_{c=1}^{C}$. Classification uses cosine similarity:
\begin{equation}
    s_c =
    \frac{f_v(x)^\top f_t(p_c)}
    {\|f_v(x)\|\,\|f_t(p_c)\|},
\end{equation}
where $p_c$ is a textual description of class $c$, such as ``a photo of a car''. Standard zero-shot classification relies on handcrafted prompt templates. Instead of handcrafted prompts, CoOp~\cite{zhou2022learning} learns $M$ context vectors $P=(v_1,\ldots,v_M)\in\mathbb{R}^{M\times d}$, constructing each class prompt as:
\begin{equation}\label{eq:coop}
    p_c=[v_1,\ldots,v_M,\text{CLASS}_c].
\end{equation}
These vectors are optimized while both encoders remain frozen. Thus, the learned prompts $P$ define the task-specific classifier parameters $\phi$, with the frozen text encoder mapping prompts to class prototypes.
\subsection{Vision-Only Backbones for Classification}
Self-supervised vision models, such as DINOv2~\cite{oquab2024dinov}, learn transferable visual representations without textual supervision. For classification, a linear head maps backbone features $\mathbf{h}=f_\theta(x)\in\mathbb{R}^{d}$ to logits:
\begin{equation}
    \mathbf{z}=W_{\mathrm{cls}}\mathbf{h},
    \qquad W_{\mathrm{cls}}\in\mathbb{R}^{C\times d}.
\end{equation}
Unlike CLIP, DINOv2 uses a directly trainable classifier, allowing us to study classifier-first training independently of prompt learning.

\subsection{Decoder-only LLMs for Classification}
Decoder-only LLMs such as Llama~3~\cite{grattafiori2024llama} can perform sequence classification using a task-specific linear head. Given tokens $\mathbf{x}=(x_1,\ldots,x_L)$, the backbone produces:
\begin{equation}
    H=f_\theta(\mathbf{x})\in\mathbb{R}^{L\times d}.
\end{equation}
We use the hidden representation $\mathbf{h}\in\mathbb{R}^{d}$ of the last non-padding token, which incorporates information from preceding tokens through causal attention. Classification then follows the same linear projection as DINOv2, $\mathbf{z}=W_{\mathrm{cls}}\mathbf{h}$, optimized using cross-entropy loss.

Despite differences in modality and classifier implementation, all three architectures follow the backbone-classifier formulation in~\cref{sec:backbone_classifier}.

\begin{figure*}[t]
    \centering
    \includegraphics[width=\textwidth]{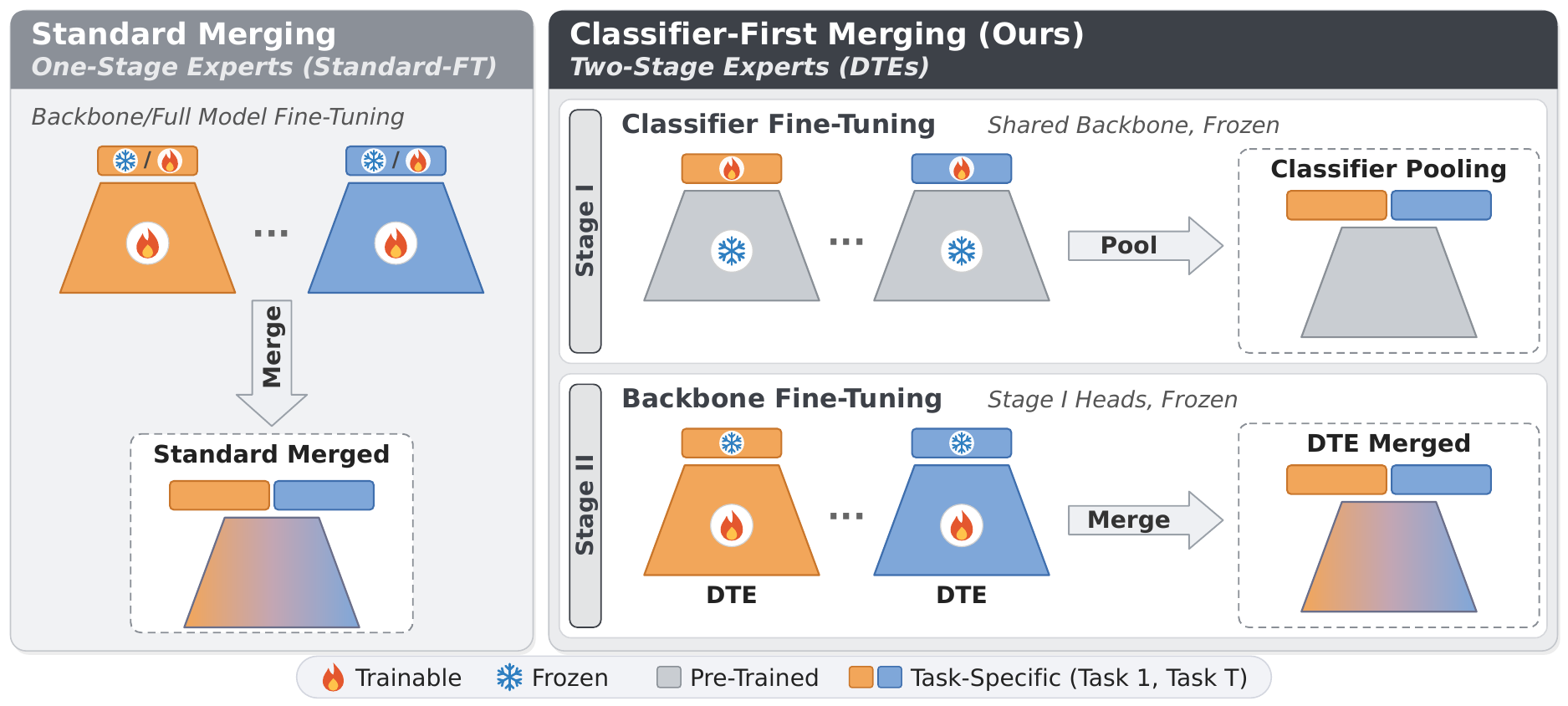}
    \caption{Method overview. \textbf{(Left)} \textbf{Standard merging}: one-stage experts are obtained by backbone or full-model fine-tuning and then merged. \textbf{(Right)} \textbf{Two-stage training of Dual-Tuned Experts (DTEs)}: in Stage I, classifier training is performed with a frozen backbone; in Stage II, the backbone is fine-tuned while keeping the learned classifier fixed.}
    \label{fig:method}
\end{figure*}
\section{Classifier-First Expert Training for Model Merging}
\label{sec:method}

\subsection{Classifier Training as a Strong Baseline for Multi-Task Inference}

We first examine classifier-only adaptation as an alternative to backbone fine-tuning. As shown in~\cref{fig:fig1long} (a), prompt learning on CLIP and classifier-only adaptation on Llama recover much of the performance of backbone/full model fine-tuning while optimizing only a small number of parameters. This suggests that part of the task adaptation can be achieved by aligning the classification interface with the representations already encoded by the pre-trained backbone.

In standard classification-based model merging, task-specific classifiers are retained separately and selected using the known task identity. Thus, training only the classifiers while keeping the backbone frozen provides a simple multi-task inference strategy that avoids weight merging altogether. As shown in~\cref{fig:fig1long} (b), this baseline is competitive with established merging methods, particularly as the number of tasks increases. However, for some tasks like SVHN, classifier-only adaptation lags significantly behind backbone fine-tuned experts, motivating subsequent adaptation of the backbone features.

\subsection{Dual-Tuned Experts (DTEs)}
We introduce \emph{Dual-Tuned Experts} (DTEs), a two-stage strategy that first trains the task-specific classifier with a frozen backbone, and then fine-tunes the backbone while keeping the trained classifier fixed. Unlike merge-time approaches, DTE modifies only expert training and is compatible with existing merging methods.

Following~\cref{sec:backbone_classifier}, we denote the pre-trained backbone parameters by $\theta_0$ and the task-specific classifier parameters by $\phi_i$. For each task $i\in\{1,\ldots,T\}$, we let $\mathcal{D}_i$ denote its training dataset and define the classification objective as:
\begin{equation}
\mathcal{L}_i(\theta,\phi_i)
=
\mathbb{E}_{(x,y)\sim\mathcal{D}_i}
\left[
\ell\left(g_{\phi_i}(f_\theta(x)),y\right)
\right],
\label{eq:dte_objective}
\end{equation}
where $\ell$ denotes the classification loss. DTE optimizes this objective in two consecutive stages.

\tit{Stage 1: Classifier Tuning.}
We first optimize the task-specific classifier while keeping the pre-trained backbone frozen:
\begin{equation}
\phi_i^{*}
=
\arg\min_{\phi_i}
\mathcal{L}_i(\theta_0,\phi_i).
\label{eq:dte_stage1}
\end{equation}
This stage adapts the classification interface to the representations provided by the pre-trained model without modifying its backbone parameters.

For CLIP, $\phi_i$ corresponds to the learnable context vectors $P_i$, with both encoders frozen. For vision-only backbones such as DINOv2, and decoder-only LLMs such as Llama~3, $\phi_i$ corresponds to a task-specific linear classification head $W_{\mathrm{cls}}^{(i)}$, optimized on top of the frozen pre-trained backbone.

\tit{Stage 2: Backbone fine-tuning.}
Once the classifier is learned, we freeze its parameters and fine-tune the backbone, initialized from the same pre-trained parameters $\theta_0$:
\begin{equation}
\theta_i^{*}
=
\arg\min_{\theta}
\mathcal{L}_i(\theta,\phi_i^{*}),
\qquad
\theta\ \text{initialized at}\ \theta_0.
\label{eq:dte_stage2}
\end{equation}

The learned classifier therefore provides a fixed task-specific prediction interface throughout backbone adaptation. Unlike conventional joint-tuning, where the classifier and backbone may change simultaneously, DTE separates the two optimization stages.

For CLIP, this corresponds to fine-tuning the vision encoder while keeping the learned prompts and text encoder fixed. For DINOv2 and decoder-only LLMs, the corresponding vision or language backbone is fine-tuned while the linear classification head learned in Stage~1 remains frozen. 

\begin{figure}[t]
    \centering
    \includegraphics[width=\columnwidth]{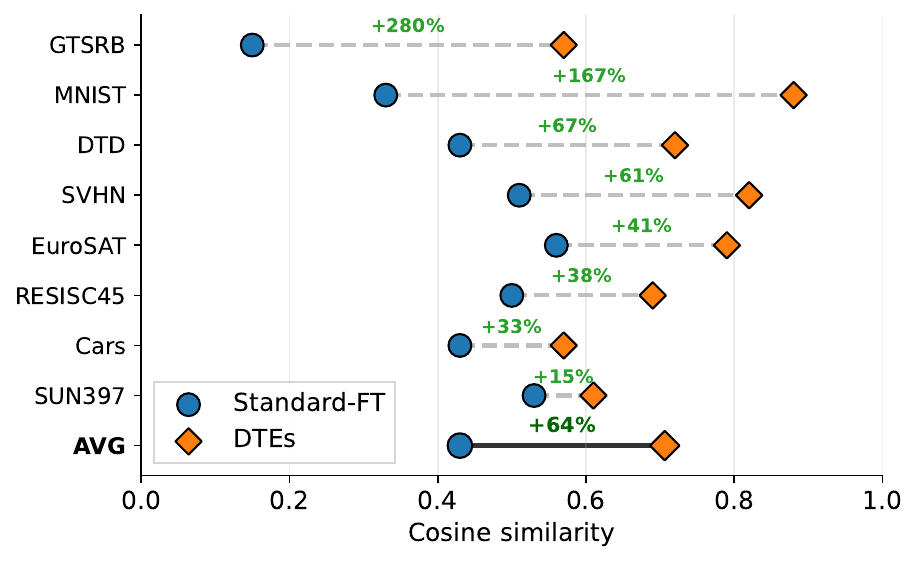}
    \caption{Cosine similarity between the pre-trained model (CLIP ViT-L/14) and the DTEs and Standard-FT experts for the 8-Vision benchmark. DTEs exhibit consistently higher similarity to the pre-trained model across all tasks, indicating smaller representation shifts after adaptation and improved compatibility for merging.}
    \label{fig:embeddings_cosine}
\end{figure}

\tit{Inference and merging.} 
Once the experts have been trained independently, we merge only their backbone parameters, following the standard model merging protocol. Defining the task vectors as $\Delta\theta_i=\theta_i^{*}-\theta_0$, the resulting shared backbone can be obtained through any compatible merging operator $\mathcal{M}$:
\begin{equation}
\theta_{\mathrm{merged}}
=
\mathcal{M}\left(
\theta_0,\{\Delta\theta_i\}_{i=1}^{T}
\right).
\label{eq:dte_merge}
\end{equation}
The learned classifiers $\{\phi_i^{*}\}_{i=1}^{T}$ are retained separately and are not merged. At inference time, the known task identity selects the corresponding classifier, yielding predictions through:
\begin{equation}
\mathbf{z}_i
=
g_{\phi_i^{*}}
\left(
f_{\theta_{\mathrm{merged}}}(x)
\right).
\label{eq:dte_inference}
\end{equation}

This preserves the standard task-aware evaluation protocol used in classification-based model merging benchmarks~\cite{ilharco2023editing,tsv,marczak2025task}. The complete training and inference procedure is illustrated in~\cref{fig:method}.

\subsection{Analysis}
\label{sec:analysis_lbl}
We hypothesize that adapting the classifier first reduces the backbone modifications required during subsequent fine-tuning. We investigate this through representation drift and interpolation losses, which provide empirical diagnostics of compatibility.

\tit{Representation drift.}
We measure how much the backbone representations change during task adaptation. Let $f_{\theta_0}$ denote the pre-trained backbone and $f_{\theta_i^{*}}$ the backbone adapted to task $i$. Given an input $x$, we define
\begin{equation}
\mathbf{h}^{\mathrm{pre}}(x)
=
f_{\theta_0}(x),
\qquad
\mathbf{h}^{(i)}(x)
=
f_{\theta_i^{*}}(x).
\end{equation}

\begin{figure}[t]
    \includegraphics[width=\columnwidth]{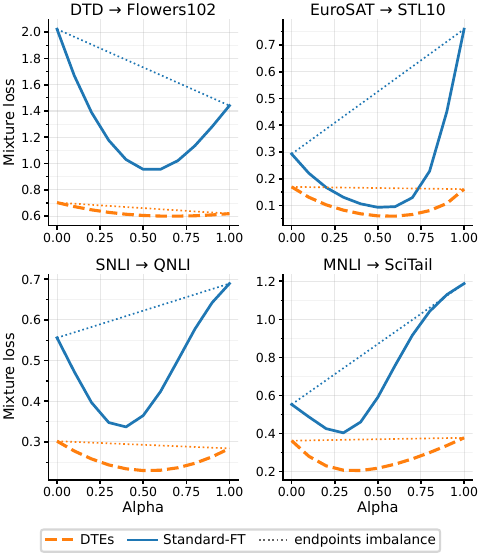}
    \caption{Mixture loss along linear interpolations between pairs of experts. Top row is CLIP ViT-B/32 and bottom row is Llama~3~8B.}
    \label{fig:loss_barrier}
\end{figure}
\begin{table*}[t]
\vspace{-1em}
	\centering
    \caption{Average Top-1 accuracy (\%) across vision tasks using CLIP ViT-B/32, ViT-L/14, and DINOv2-B/14 with normalized accuracy relative to the corresponding experts in brackets.}
	\label{tab:compare_base_ours}
	\resizebox{\linewidth}{!}{
		\begin{tabular}{l c ccc c ccc c ccc}
			\toprule
			& & \multicolumn{3}{c}{\textbf{ViT-B/32}} 
			&& \multicolumn{3}{c}{\textbf{ViT-L/14}}
			&& \multicolumn{3}{c}{\textbf{DINOv2-B/14}} \\
			\cmidrule(lr){3-5} \cmidrule(lr){7-9} \cmidrule(lr){11-13}
			\textbf{Method} & \textbf{DTE} & \textbf{8 Tasks} & \textbf{14 Tasks} & \textbf{20 Tasks} && \textbf{8 Tasks} & \textbf{14 Tasks} & \textbf{20 Tasks} && \textbf{8 Tasks} & \textbf{14 Tasks} & \textbf{20 Tasks} \\
			\midrule
			Zero-shot & -- & 48.24 & 56.43 & 55.56 && 65.25 & 67.98 & 65.04 && -- & -- & -- \\
			\midrule
			Standard-FT & -- & 90.28$_{\scriptstyle(100.0)}$ & 89.31$_{\scriptstyle(100.0)}$ & 90.26$_{\scriptstyle(100.0)}$ && 93.91$_{\scriptstyle(100.0)}$ & 93.10$_{\scriptstyle(100.0)}$ & 93.87$_{\scriptstyle(100.0)}$ && 91.35$_{\scriptstyle(100.0)}$ & 91.08$_{\scriptstyle(100.0)}$ & 90.78$_{\scriptstyle(100.0)}$ \\
			DTEs & -- & 90.13$_{\scriptstyle(100.0)}$ & 89.54$_{\scriptstyle(100.0)}$ & 90.52$_{\scriptstyle(100.0)}$ && 93.61$_{\scriptstyle(100.0)}$ & 93.13$_{\scriptstyle(100.0)}$ & 93.87$_{\scriptstyle(100.0)}$ && 92.97$_{\scriptstyle(100.0)}$ & 92.41$_{\scriptstyle(100.0)}$ & 91.89$_{\scriptstyle(100.0)}$ \\
			\midrule
			Classifier-Only & -- & 78.96 & 81.01 & 81.72 && 87.28 & 87.95 & 87.92 && 85.33 & 87.18 & 87.20 \\
			\midrule
			\multirow{2}{*}{WA}
			& \redxmark & 65.78$_{\scriptstyle(72.9)}$ & 63.51$_{\scriptstyle(71.1)}$ & 60.81$_{\scriptstyle(67.4)}$  && 79.21$_{\scriptstyle(84.4)}$ & 76.52$_{\scriptstyle(82.2)}$ & 71.39$_{\scriptstyle(76.1)}$ && 29.53$_{\scriptstyle(32.3)}$ & 21.17$_{\scriptstyle(23.2)}$ & 24.24$_{\scriptstyle(26.7)}$ \\
			& \greencmark & 82.47$_{\scriptstyle(91.5)}$ & 81.55$_{\scriptstyle(91.1)}$ & 81.17$_{\scriptstyle(89.7)}$ && 89.59$_{\scriptstyle(95.7)}$ & 89.07$_{\scriptstyle(95.6)}$ & 88.69$_{\scriptstyle(94.5)}$ && 87.63$_{\scriptstyle(94.3)}$ & 87.92$_{\scriptstyle(95.1)}$ & 87.14$_{\scriptstyle(94.8)}$ \\
			\midrule
			\multirow{2}{*}{TA}
			& \redxmark
			& 70.08$_{\scriptstyle(77.6)}$
			& 64.92$_{\scriptstyle(72.7)}$
			& 60.81$_{\scriptstyle(67.4)}$
			&& 84.16$_{\scriptstyle(89.6)}$
			& 79.78$_{\scriptstyle(85.7)}$
			& 73.65$_{\scriptstyle(78.5)}$
			&& 45.15$_{\scriptstyle(49.4)}$ & 37.40$_{\scriptstyle(41.1)}$ & 33.05$_{\scriptstyle(36.4)}$ \\
			& \greencmark
			& 82.47$_{\scriptstyle(91.5)}$
			& 81.81$_{\scriptstyle(91.4)}$
			& 81.17$_{\scriptstyle(89.7)}$
			&& 89.61$_{\scriptstyle(95.7)}$
			& 89.07$_{\scriptstyle(95.6)}$
			& 88.69$_{\scriptstyle(94.5)}$
			&& 87.63$_{\scriptstyle(94.3)}$ & 87.92$_{\scriptstyle(95.1)}$ & 87.20$_{\scriptstyle(94.9)}$ \\
			\midrule
			
			\multirow{2}{*}{TIES}
			& \redxmark
			& 73.40$_{\scriptstyle(81.3)}$
			& 65.04$_{\scriptstyle(72.8)}$
			& 62.38$_{\scriptstyle(69.1)}$
			&& 85.62$_{\scriptstyle(91.2)}$
			& 78.27$_{\scriptstyle(84.1)}$
			& 74.24$_{\scriptstyle(79.1)}$
			&& 52.44$_{\scriptstyle(57.4)}$ & 40.00$_{\scriptstyle(43.9)}$ & 40.13$_{\scriptstyle(44.2)}$ \\
			& \greencmark
			& 83.05$_{\scriptstyle(92.1)}$
			& 81.80$_{\scriptstyle(91.4)}$
			& 82.18$_{\scriptstyle(90.8)}$
			&& 90.29$_{\scriptstyle(96.4)}$
			& 89.01$_{\scriptstyle(95.6)}$
			& 88.73$_{\scriptstyle(94.5)}$
			&& 88.96$_{\scriptstyle(95.7)}$ & 88.52$_{\scriptstyle(95.8)}$ & 87.20$_{\scriptstyle(94.9)}$ \\
			\midrule
			
			\multirow{2}{*}{TSV-M}
			& \redxmark
			& 83.25$_{\scriptstyle(92.2)}$
			& 79.60$_{\scriptstyle(89.1)}$
			& 76.73$_{\scriptstyle(85.0)}$
			&& 91.00$_{\scriptstyle(96.9)}$
			& 88.35$_{\scriptstyle(94.9)}$
			& 86.35$_{\scriptstyle(92.0)}$
			&& 74.88$_{\scriptstyle(82.0)}$ & 69.51$_{\scriptstyle(76.3)}$ & 63.58$_{\scriptstyle(70.0)}$ \\
			& \greencmark
			& 86.70$_{\scriptstyle(96.2)}$
			& 84.57$_{\scriptstyle(94.5)}$
			& 84.71$_{\scriptstyle(93.6)}$
			&& 91.75$_{\scriptstyle(98.0)}$
			& 90.47$_{\scriptstyle(97.1)}$
			& 89.96$_{\scriptstyle(95.8)}$
			&& 90.86$_{\scriptstyle(97.7)}$ & 89.83$_{\scriptstyle(97.2)}$ & 88.75$_{\scriptstyle(96.6)}$ \\
			\midrule
			
			\multirow{2}{*}{Iso-C}
			& \redxmark
			& 83.67$_{\scriptstyle(92.7)}$
			& 78.90$_{\scriptstyle(88.3)}$
			& 74.11$_{\scriptstyle(82.1)}$
			&& 92.27$_{\scriptstyle(98.3)}$
			& 88.33$_{\scriptstyle(94.9)}$
			& 85.35$_{\scriptstyle(90.9)}$
			&& 80.08$_{\scriptstyle(87.7)}$ & 72.17$_{\scriptstyle(79.2)}$ & 63.90$_{\scriptstyle(70.4)}$ \\
			& \greencmark
			& 86.59$_{\scriptstyle(96.1)}$
			& 84.18$_{\scriptstyle(94.0)}$
			& 83.97$_{\scriptstyle(92.8)}$
			&& 92.39$_{\scriptstyle(98.7)}$
			& 90.55$_{\scriptstyle(97.2)}$
			& 90.40$_{\scriptstyle(96.3)}$
			&& 91.20$_{\scriptstyle(98.1)}$ & 89.80$_{\scriptstyle(97.2)}$ & 88.36$_{\scriptstyle(96.2)}$ \\
			\bottomrule
	\end{tabular}}
\end{table*}
We quantify representation preservation using the average cosine similarity between pre-trained and adapted features:
\begin{equation}
\mathrm{sim}_i
=
\mathbb{E}_{x\sim\mathcal{D}_i^{\mathrm{test}}}
\left[
\frac{
f_{\theta_0}(x)^\top f_{\theta_i^{*}}(x)
}{
\|f_{\theta_0}(x)\|
\|f_{\theta_i^{*}}(x)\|
}
\right].
\label{eq:representation_drift}
\end{equation}
For CLIP, we compare image embeddings; for Llama~3, we use the hidden representation of the last non-padding token.

As shown in~\cref{fig:embeddings_cosine} for CLIP, DTEs consistently preserve representations closer to the pre-trained backbone. Average cosine similarity improves by approximately $64\%$ on the 8-task CLIP benchmark and $43\%$ across six Llama~3 classification tasks, relative to standard fine-tuning. Llama~3 results are provided in the supplementary material.

These findings suggest that, across different model architectures, adapting the classification interface before the backbone helps preserve pre-trained representations, supporting our hypothesis that classifier-first training reduces unnecessary changes to backbone features.

We further measure the similarity between merged and independently trained expert representations on CLIP. Merged DTEs achieve approximately $59\%$ higher relative cosine similarity than standard fine-tuned experts, indicating better preservation of task-specific representations. Details are provided in the supplementary material.

\tit{Linear interpolation.}
We further examine the loss landscape along linear interpolations between task-specific backbones. Given experts trained on tasks $i$ and $j$, we define:
\begin{equation}
\theta(\alpha)
=
(1-\alpha)\theta_i^{*}
+
\alpha\theta_j^{*},
\qquad
\alpha\in[0,1].
\label{eq:dte_interpolation}
\end{equation}
The task-specific classifiers remain fixed throughout interpolation. We evaluate the mixture loss over tasks $i$ and $j$:
\begin{equation}
\mathcal{L}_{\mathrm{mix}}(\alpha)
=
\frac{1}{2}
\left[
\mathcal{L}_i(\theta(\alpha),\phi_i)
+
\mathcal{L}_j(\theta(\alpha),\phi_j)
\right],
\label{eq:dte_mixture_loss}
\end{equation}
where the classifiers correspond to the evaluated expert family.

As shown in~\cref{fig:loss_barrier}, DTEs exhibit more favorable interpolation profiles than standard fine-tuned experts across the evaluated CLIP and Llama~3 task pairs. 

These findings are consistent with recent work connecting geometric properties of task updates to downstream mergeability~\cite{zhou2026demystifying}. Together, they support the empirical benefits of classifier-first training observed across architectures.

\section{Experimental Results}
\label{sec:exp}
\tit{Experimental setup.}
We compare merged Dual-Tuned Experts (DTEs) against conventional one-stage experts (Standard-FT) using Weight Averaging (WA)~\cite{wortsman2022model}, Task Arithmetic (TA)~\cite{ilharco2023editing}, TSV-M~\cite{tsv}, Iso-C~\cite{marczak2025task}, and TIES-Merging~\cite{yadav2023tiesmerging}.

We test CLIP and DINOv2 on 8, 14, and 20 vision tasks following~\cite{ilharco2023editing,tsv}, as well as Llama~3 on 6 NLI tasks following~\cite{stoica2024knots,panariello2025accurate}. Further details on methods and datasets in the supplementary material.

\tit{Implementation details.} For vision experiments, we use the publicly available OpenAI CLIP ViT-B/32 and ViT-L/14 checkpoints~\cite{radford2021learning}, and the official Meta DINOv2 ViT-B/14 checkpoint~\cite{oquab2024dinov}. Both CLIP and DINOv2 follow the same backbone fine-tuning protocol as Task Arithmetic~\cite{ilharco2023editing}. For CLIP, classifier adaptation follows CoOp's unified-context formulation~\cite{zhou2022learning} with $M=16$ learnable tokens, while DINOv2 uses a task-specific linear classification head. For language experiments, we use the publicly available Meta Llama~3 8B checkpoint~\cite{grattafiori2024llama}, following the training protocol of KnOTS~\cite{stoica2024knots}.

Within each architecture, Standard-FT and Stage~2 of DTE use the same optimizer, learning rate, training steps, and data splits. Stage~1 uses learning rates of \num{1e-3} for CLIP prompts and DINOv2 classifiers, and \num{3e-4} for Llama classification heads.

For merging methods with a scaling coefficient, we select $\lambda$ using a held-out validation set, following prior evaluation protocols~\cite{ilharco2023editing,yadav2023tiesmerging, tsv,panariello2025accurate,marczak2025task}.

\tit{Evaluation interfaces.} Each expert family is evaluated with its retained task-specific classifiers. For CLIP, Standard-FT uses handcrafted text-derived classifiers, whereas DTEs use learned prompts. For DINOv2 and Llama, the corresponding trained linear heads are retained. These comparisons measure the performance of complete merged systems; they do not isolate the contribution of backbone changes from that of classifier changes.

\subsection{Merging Vision Experts}
\label{sec:mergevision}
We first evaluate full-parameter adaptation using CLIP ViT-B/32, CLIP ViT-L/14, and DINOv2-B/14. \Cref{tab:compare_base_ours} reports standalone expert accuracy, classifier-only accuracy, and accuracy after merging.

\tit{CLIP.} Across both CLIP architectures and increasing numbers of tasks, DTE consistently improves merged-system accuracy under the evaluated merging operators. Importantly, these improvements arise despite comparable standalone expert performance, indicating that they cannot be explained solely by differences in individual expert accuracy. DTEs gains are particularly pronounced for linear merging methods.

\tit{DINOv2.}
To assess whether classifier-first training generalizes beyond CLIP's text-based classification interface, we evaluate DINOv2 with task-specific linear classification heads across 8, 14, and 20 tasks.

Despite modest improvements in standalone expert accuracy, DTE substantially improves downstream merging performance. For instance, on the 8-task benchmark, WA improves from $29.53\%$ to $87.63\%$, while Iso-C increases from $80.08\%$ to $91.20\%$, retaining $98.1\%$ of the corresponding standalone expert accuracy.

Importantly, these improvements persist as the number of tasks increases. On 14 tasks, DTE improves TSV-M from $69.51\%$ to $89.83\%$ and Iso-C from $72.17\%$ to $89.80\%$. On 20 tasks, the corresponding improvements are from $63.58\%$ to $88.75\%$ and from $63.90\%$ to $88.36\%$, respectively. While conventional merging degrades substantially with additional tasks, DTE maintains comparatively stable performance, retaining $96.6\%$ of standalone expert accuracy with TSV-M on 20 tasks.

Classifier-only adaptation also provides a strong baseline, reaching $87.18\%$ and $87.20\%$ on 14 and 20 tasks, respectively. Nevertheless, DTE with TSV-M improves upon this baseline by $2.65$ and $1.55$ percentage points, while enabling additional backbone adaptation. These results show that classifier-first training improves the performance of merged systems beyond CLIP, without relying on extra parameters.

\tit{Equal computational budget.}
We additionally match DTE's total classifier and backbone training budget to Standard-FT. Merging performance remains very close to the unconstrained DTE setting, suggesting that the gains do not require additional training compute. Details are provided in the supplementary material.

\tit{LoRA adaptation.}
We further evaluate CLIP ViT-B/32 using LoRA~\cite{hu2021lora} ($r=16$), merging the learned updates in both full parameter space and Core Space~\cite{panariello2025accurate}.

\begin{table}[t]
\centering
\caption{Average Top-1 accuracy (\%) across vision tasks with normalized accuracy relative to the corresponding experts in brackets.}
\label{tab:lora_table}
\setlength{\tabcolsep}{3.8pt}
\resizebox{\linewidth}{!}{
\begin{tabular}{lccccc}
\toprule
&&&\multicolumn{3}{c}{\textbf{LoRA ViT-B/32}}\\
\cmidrule(lr){4-6}
\textbf{Method} & \textbf{Subspace} & \textbf{DTE} & \textbf{8 Tasks} & \textbf{14 Tasks} & \textbf{20 Tasks} \\
\midrule
Standard-FT
& -- & -- & 88.59$_{\scriptstyle(100.0)}$ & 87.96$_{\scriptstyle(100.0)}$ & 89.06$_{\scriptstyle(100.0)}$ \\

DTEs
& -- & -- & 87.73$_{\scriptstyle(100.0)}$ & 87.19$_{\scriptstyle(100.0)}$ & 87.73$_{\scriptstyle(100.0)}$ \\
\midrule

Classifier-Only & -- & -- & 78.96 & 81.01 & 81.72 \\
\midrule
\multirow{2}{*}{WA}
& -- & \redxmark & 72.98$_{\scriptstyle(82.38)}$ & 62.73$_{\scriptstyle(71.32)}$ & 61.64$_{\scriptstyle(69.21)}$ \\
& -- & \greencmark & 83.62$_{\scriptstyle(95.32)}$ & 82.77$_{\scriptstyle(94.93)}$ & 81.89$_{\scriptstyle(93.34)}$ \\
\midrule

\multirow{2}{*}{TA}
& -- & \redxmark & 74.52$_{\scriptstyle(84.12)}$ & 69.26$_{\scriptstyle(78.74)}$ & 67.82$_{\scriptstyle(76.15)}$ \\
& -- & \greencmark & 85.04$_{\scriptstyle(96.93)}$ & 83.14$_{\scriptstyle(95.35)}$ & 83.21$_{\scriptstyle(94.85)}$ \\
\midrule

\multirow{4}{*}{TIES}
& \multirow{2}{*}{Full} & \redxmark & 73.20$_{\scriptstyle(82.63)}$ & 71.93$_{\scriptstyle(81.78)}$ & 70.09$_{\scriptstyle(78.70)}$ \\
&  & \greencmark & 84.94$_{\scriptstyle(96.82)}$ & 83.92$_{\scriptstyle(96.25)}$ & 83.63$_{\scriptstyle(95.33)}$ \\
\cmidrule(lr){2-6}
& \multirow{2}{*}{Core} & \redxmark & 74.95$_{\scriptstyle(84.60)}$ & 73.67$_{\scriptstyle(83.75)}$ & 71.28$_{\scriptstyle(80.04)}$ \\
&  & \greencmark & 84.46$_{\scriptstyle(96.27)}$ & 81.64$_{\scriptstyle(93.63)}$ & 83.18$_{\scriptstyle(94.81)}$ \\
\midrule

\multirow{4}{*}{TSV-M}
& \multirow{2}{*}{Full} & \redxmark & 75.93$_{\scriptstyle(85.71)}$ & 75.78$_{\scriptstyle(86.15)}$ & 74.03$_{\scriptstyle(83.12)}$ \\
&  & \greencmark & 85.62$_{\scriptstyle(97.59)}$ & 84.67$_{\scriptstyle(97.11)}$ & 84.89$_{\scriptstyle(96.76)}$ \\
\cmidrule(lr){2-6}
& \multirow{2}{*}{Core} & \redxmark & 78.90$_{\scriptstyle(89.06)}$ & 78.25$_{\scriptstyle(88.96)}$ & 75.56$_{\scriptstyle(84.84)}$ \\
&  & \greencmark & 86.45$_{\scriptstyle(98.54)}$ & 85.18$_{\scriptstyle(97.69)}$ & 85.60$_{\scriptstyle(97.57)}$ \\
\midrule

\multirow{4}{*}{Iso-C}
& \multirow{2}{*}{Full} & \redxmark & 69.36$_{\scriptstyle(78.29)}$ & 69.93$_{\scriptstyle(79.50)}$ & 69.92$_{\scriptstyle(78.51)}$ \\
&  & \greencmark & 80.51$_{\scriptstyle(91.77)}$ & 81.89$_{\scriptstyle(93.92)}$ & 83.07$_{\scriptstyle(94.69)}$ \\
\cmidrule(lr){2-6}
& \multirow{2}{*}{Core} & \redxmark & 74.87$_{\scriptstyle(84.51)}$ & 71.33$_{\scriptstyle(81.09)}$ & 71.02$_{\scriptstyle(79.74)}$ \\
&  & \greencmark & 86.05$_{\scriptstyle(98.09)}$ & 84.57$_{\scriptstyle(97.00)}$ & 84.49$_{\scriptstyle(96.31)}$ \\
\bottomrule
\end{tabular}}
\end{table}

As shown in~\cref{tab:lora_table}, standard LoRA merging suffers substantial degradation in full space~\cite{stoica2024knots,panariello2025accurate}, whereas DTE considerably reduces this effect. On 20 tasks, even standard TA improves from $67.82\%$ to $83.21\%$. Although TSV-M and Iso-C provide further improvements, the performance gap between simple and advanced methods becomes smaller with DTE.

\begin{table}[t]
	\centering
	\small
    \caption{Joint-task setting absolute accuracy (\%) of merged models on the 8-vision tasks with ViT-B/32.}
	\label{tab:joint-task}
	\resizebox{\columnwidth}{!}{%
	\begin{tabular}{lrrrrr}
		\toprule
		Method & WA & TA & TIES & TSV-M & Iso-C \\
		\midrule
		Standard-FT
		& 60.55 & 63.59 & 66.95 & 77.16 & 78.29 \\
		DTEs
		& \textbf{73.83} & \textbf{74.41} & \textbf{75.78} & \textbf{83.92} & \textbf{82.69} \\
		\bottomrule
	\end{tabular}%
	}
\end{table}

\tit{Joint-task evaluation.}
We additionally evaluate a more challenging joint-task setting in which the task identity is unavailable. Following~\cite{stoica2024knots}, we construct a shared classification interface over the union of classes from the eight CLIP vision tasks. This requires the merged model to distinguish between classes across tasks rather than selecting a task-specific classifier. Further details provided in the supplementary material.

As shown in~\cref{tab:joint-task}, DTE improves accuracy over Standard-FT across all evaluated merging methods. These results indicate that the benefits of classifier-first training are not restricted to inference with an explicitly provided task identity.

\begin{table}[t]
\centering
\caption{Average Top-1 accuracy (\%) across 6 NLI tasks with normalized accuracy relative to the corresponding experts in brackets.}
\label{tab:lora_llama_table}
\setlength{\tabcolsep}{8pt}
\resizebox{.9\linewidth}{!}{
\begin{tabular}{lccc}
\toprule
&&&\textbf{LoRA Llama 3 8B}\\
\cmidrule(lr){4-4}
\textbf{Method} & \textbf{Subspace} & \textbf{DTE} & \textbf{6 NLI Tasks} \\
\midrule
Standard-FT
& -- & -- & 92.48$_{\scriptstyle(100.0)}$ \\

DTEs
& -- & -- & 90.89$_{\scriptstyle(100.0)}$ \\
\midrule

Classifier-Only & -- & -- & 78.67 \\
\midrule
\multirow{2}{*}{WA}
& -- & \redxmark & 59.55$_{\scriptstyle(64.39)}$ \\
& -- & \greencmark & 86.47$_{\scriptstyle(95.14)}$ \\
\midrule

\multirow{2}{*}{TA}
& -- & \redxmark & 69.30$_{\scriptstyle(74.93)}$ \\
& -- & \greencmark & 89.00$_{\scriptstyle(97.92)}$ \\
\midrule

\multirow{4}{*}{TIES}
& \multirow{2}{*}{Full} & \redxmark & 69.80$_{\scriptstyle(75.47)}$ \\
&  & \greencmark & 89.03$_{\scriptstyle(97.95)}$ \\
\cmidrule(lr){2-4}
& \multirow{2}{*}{Core} & \redxmark & 75.48$_{\scriptstyle(81.62)}$ \\
&  & \greencmark & 87.79$_{\scriptstyle(96.58)}$ \\
\midrule

\multirow{4}{*}{TSV-M}
& \multirow{2}{*}{Full} & \redxmark & 77.55$_{\scriptstyle(83.85)}$ \\
&  & \greencmark & 88.35$_{\scriptstyle(97.20)}$ \\
\cmidrule(lr){2-4}
& \multirow{2}{*}{Core} & \redxmark & 79.63$_{\scriptstyle(86.10)}$ \\
&  & \greencmark & 89.10$_{\scriptstyle(98.03)}$ \\
\midrule

\multirow{4}{*}{Iso-C}
& \multirow{2}{*}{Full} & \redxmark & 48.41$_{\scriptstyle(52.34)}$ \\
&  & \greencmark & 78.91$_{\scriptstyle(86.82)}$ \\
\cmidrule(lr){2-4}
& \multirow{2}{*}{Core} & \redxmark & 77.52$_{\scriptstyle(83.82)}$ \\
&  & \greencmark & 88.49$_{\scriptstyle(97.36)}$ \\
\bottomrule
\end{tabular}}
\end{table}

\subsection{Merging Language Experts}
We evaluate whether classifier-first training extends to decoder-only language models using Llama~3 8B. Following~\cite{stoica2024knots}, we independently adapt the model to six NLI tasks using LoRA ($r=16$) applied to each attention layer. The resulting updates are merged in both full parameter space and Core Space~\cite{panariello2025accurate}.

\Cref{tab:lora_llama_table} compares the resulting merged systems. As in the vision experiments, DTE improves performance while maintaining comparable standalone expert accuracy. These results extend the benefits of classifier-first adaptation to language representations and linear sequence-classification heads, without modifying the merging algorithms.

\subsection{Merging Heterogeneous Models}
We investigate whether DTEs can be combined with Standard-FT experts within the same merged model. On the 20-task CLIP benchmark, we progressively replace Standard-FT experts with their DTE counterparts, starting from an all-Standard-FT collection and ending with an all-DTE collection. Tasks are selected in random order, and results are averaged over three runs.

\begin{figure}[t]
    \centering
    \includegraphics[width=\columnwidth]{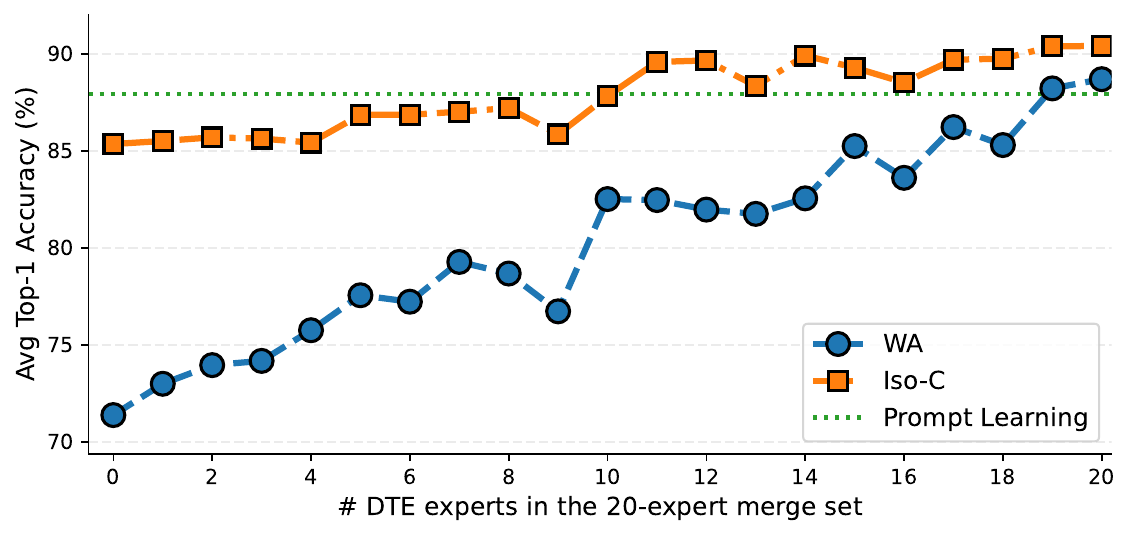}
    \caption{Performance as Standard-FT experts are progressively replaced by DTEs in a heterogeneous 20-task merge setting.}
    \label{fig:heterogeneous_merging}
    \vspace{-.9em}
\end{figure}

To avoid the influence of different scaling coefficients and to maintain a fair comparison, we report weight-averaging (WA) results throughout the entire process. In addition, we report Iso-C results to evaluate an advanced merging method on these mixed sets. Both approaches use the benchmark's standard task-specific class interface. As shown in~\cref{fig:heterogeneous_merging}, merged-system performance increases as DTEs progressively replace Standard-FT experts, under both WA and Iso-C. The experiment uses each expert's native classification interface, demonstrating that independently trained expert families can be combined within the same task-aware system.

\section{Conclusion and Discussion}
\label{sec:disc}
In this work, we investigated how classifier-first adaptation affects the training of experts for model merging. We first showed that classifier-only adaptation provides a competitive multi-task baseline without requiring backbone merging, although further backbone adaptation remains beneficial. Building on this observation, we introduced \emph{Dual-Tuned Experts} (DTEs), which first optimize task-specific classifiers and subsequently fine-tune the backbone while keeping these classifiers fixed.

Across CLIP, DINOv2, and Llama~3, DTEs improve merged-system performance, under both full-parameter and low-rank adaptation. These gains are substantially larger than differences in standalone expert accuracy, highlighting the importance of the expert-training procedure for downstream merging. Our representation and interpolation analyses further suggest that classifier-first training better preserves pre-trained features and produces more favorable interpolation profiles. Together, these findings establish classifier-first adaptation as a complementary direction to merge-time methods and merge-aware optimization.

\tinytit{Limitations.} Most experiments follow a task-aware evaluation protocol requiring the task identity to select the appropriate classifier, although our joint-task CLIP experiments relax this assumption. DTE also introduces a separate classifier-training stage and requires storing task-specific classifiers. Finally, because standard and DTE systems may use different learned classification interfaces, improved end-to-end merging performance does not necessarily imply intrinsically more compatible backbone updates. Further controlled comparisons and combinations with merge-aware training methods remain promising directions.

\section*{LLM Usage Statement}
We used Claude Opus~5 (Anthropic) in two ways. (i) \emph{Writing:} for grammar correction and to rewrite and refine parts of the manuscript, including visuals; all text and visuals were reviewed and edited by the authors, who take full responsibility for its content. (ii) \emph{Code:} as an assistant during the development of the training, inference, and evaluation code; all LLM-assisted code was reviewed and tested by the authors, and all reported results were produced
and verified by the authors.

\section*{Acknowledgements}
We acknowledge Grant PID2025-175937NB-I00 and Grant AIA2025-163919-C52 funded by MICIU/AEI/10.13039/501100011033 and the European Union’s Horizon Europe research and innovation programme under grant agreement number 101214398 (ELLIOT). The work was partially funded by Villanova, a project financed by IPCEI CIS, Prog. n. SA. 102519 - CUP B29J24000850005.
{
    \small
    \bibliographystyle{ieeenat_fullname}
    \bibliography{main}
}

\clearpage
\maketitlesupplementary
\appendix
\section{Methods and Datasets}
\label{sec:metdata}
\subsection{Methods}

We evaluate our approach with several representative model merging methods commonly used in the literature.

\noindent \textbf{Weight Averaging (WA)}~\cite{wortsman2022model}. 
Weight averaging simply computes the average of expert parameters. 
Given experts with parameters $W_i$, the merged model is obtained as:
\begin{equation}
W_{\text{merged}} = \frac{1}{T}\sum_{i=1}^{T} W_i .
\end{equation}
Despite its simplicity, weight averaging can perform well when experts lie in the same basin of the loss landscape.

\noindent \textbf{Task Arithmetic (TA)}~\cite{ilharco2023editing}. 
Task Arithmetic represents each expert as a task vector $\Delta W_i = W_i - W_0$ with respect to the pre-trained model $W_0$. 
Experts are merged through a linear combination of task vectors:
\begin{equation}
W_{\text{merged}} = W_0 + \lambda \sum_{i=1}^{T} \Delta W_i ,
\end{equation}
where $\lambda$ is a scaling coefficient selected on a validation set.

\noindent \textbf{TIES-Merging}~\cite{yadav2023tiesmerging}. 
TIES reduces destructive interference between task vectors by enforcing sign consistency across parameters. 
Conflicting parameters are removed through pruning, and the remaining updates are combined using a majority-sign rule.

\noindent \textbf{TSV-M}~\cite{tsv}. 
Task Singular Vectors (TSV) performs merging in a low-dimensional subspace obtained through singular value decomposition of task vectors. 
By projecting updates into a shared subspace, TSV reduces interference between tasks and improves merging stability.

\noindent \textbf{Iso-C}~\cite{marczak2025task}. 
Iso-C first sums up task vectors per layer and then decomposes them using singular value decomposition. 
The singular directions of the summed update are retained, while the singular values are replaced with the average singular values of the individual task vectors. 
The merged update is then reconstructed from this modified spectrum, producing a more balanced combination of task updates.

\subsection{Datasets}
For vision, the 8-task benchmark~\cite{ilharco2023editing} consists of Cars~\cite{cars}, DTD~\cite{dtd}, EuroSAT~\cite{helber_eurosat_2019}, GTSRB~\cite{stallkamp2012man}, MNIST~\cite{lecun2010mnist}, RESISC45~\cite{cheng2017remote}, SUN397~\cite{sun397}, and SVHN~\cite{svhn}. 
The 14-task benchmark~\cite{tsv} extends this set with CIFAR100~\cite{krizhevsky2009learning}, STL10~\cite{coates_analysis_2011}, Flowers102~\cite{nilsback_automated_2008}, Oxford-IIITPet~\cite{parkhi12a}, PCAM~\cite{10.1007/978-3-030-00934-2_24}, and FER2013~\cite{fer}. 
Finally, the 20-task benchmark~\cite{tsv} further includes EMNIST~\cite{cohen_emnist_2017}, CIFAR10~\cite{krizhevsky2009learning}, Food101~\cite{bossard_food-101_2014}, FashionMNIST~\cite{xiao2017/online}, RenderedSST2~\cite{Socher2013RecursiveDM}, and KMNIST~\cite{clanuwat_deep_2018}. For language, the 6 NLI tasks benchmark~\cite{panariello2025accurate,stoica2024knots} consists of SNLI~\cite{bowman2015large}, MNLI~\cite{williams2017broad1coverage}, SICK~\cite{marelli2014sick}, QNLI~\cite{wang2018glue}, RTE~\cite{wang2018glue}, and SciTail~\cite{scitail}. 

\section{Additional Analysis and Ablations}
\label{sec:addresults}
\subsection{Per-Task Performance Analysis}
\begin{figure*}[t]
    \centering
    \includegraphics[width=\textwidth]{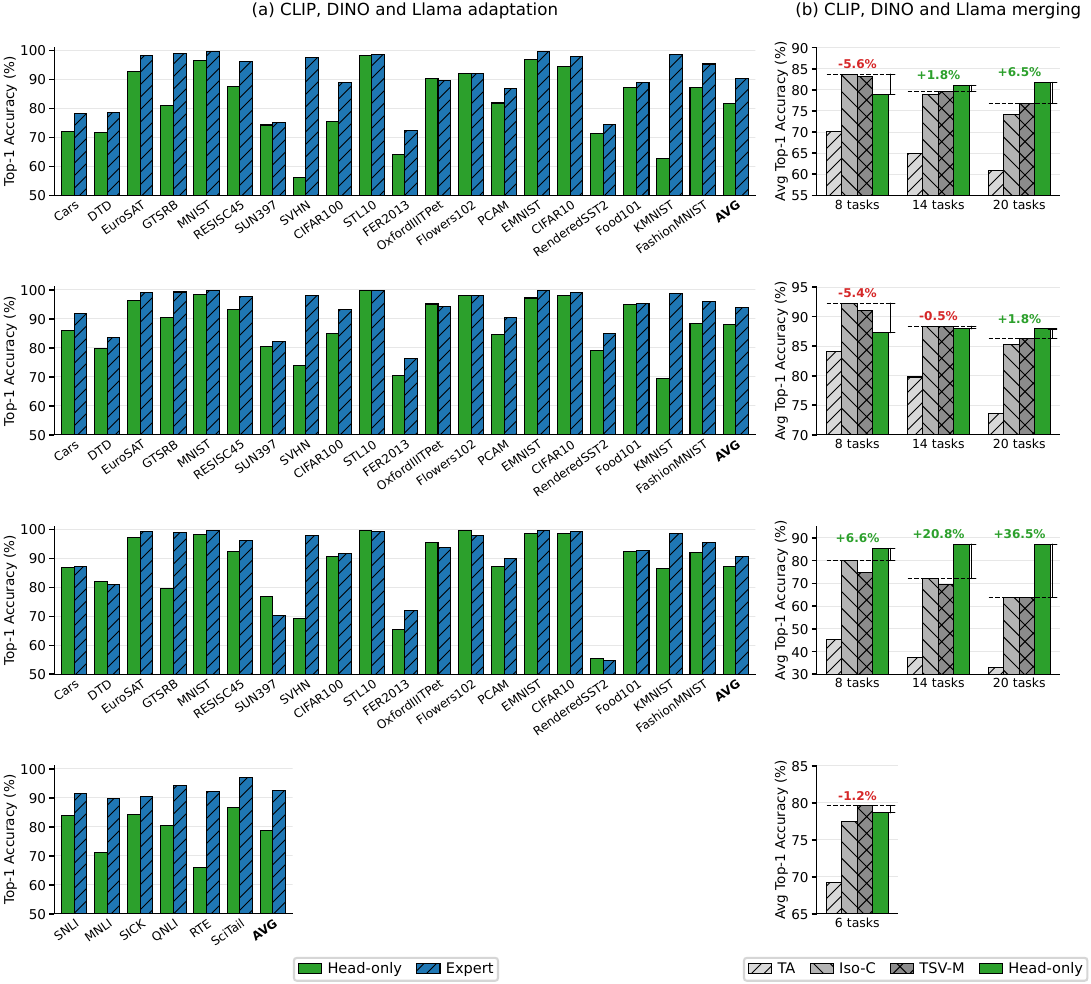}
    \caption{\textbf{(a)} Performance of CLIP ViT-B/32, CLIP ViT-L/14, DINOv2-B/14 on the 20-vision tasks and Llama~3 8B on 6 NLI tasks. Results include classifier-only and full model (Llama, DINO) or backbone (CLIP) training. \textbf{(b)} Average performance of merging CLIP ViT-B/32, CLIP ViT-L/14, DINOv2-B/14 across 8, 14, and 20 vision tasks and Llama~3 8B across 6 NLI tasks. Results compare Task Arithmetic (TA), Iso-C, and TSV-M against a classifier-only training baseline.
    }
    \label{fig:plvsvsft}
\end{figure*}

\minisection{Classifier-Only vs. Standard-FT.}
Here we also provide results for the per-task performance analysis, comparing Classifier-Only with Standard-FT for all models and tasks. As shown in \cref{fig:plvsvsft}, although Classifier-only adaptation serves as a strong alternative to Standard-FT and a strong baseline for merging, it exhibits a notable performance drop on certain datasets, particularly SVHN and KMNIST for vision. For SVHN, we hypothesize that the presence of nearby digits and cluttered backgrounds requires the backbone to learn to suppress distracting context and focus on the target digit. In the case of KMNIST, the cursive Japanese characters are largely absent from CLIP’s pretraining data, leaving the frozen backbone without meaningful visual representations for this domain. Consequently, subsequent backbone fine-tuning, like proposed by our DTEs, becomes necessary in these cases, to enable the model to learn the required task features.

\begin{figure*}[t]
    \centering
    \includegraphics[width=0.98\textwidth]{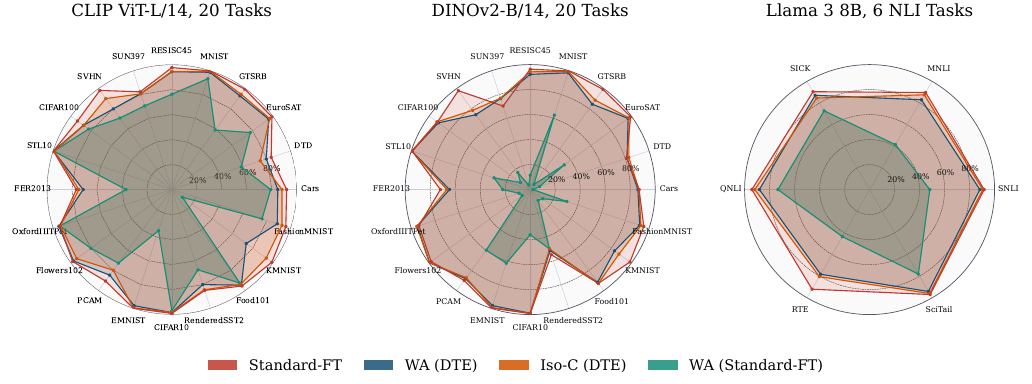}
    \caption{Per-Task Accuracy (\%) comparison of Standard-FT, WA (DTE), Iso-C (DTE) and WA (Standard-FT) for the 20-Vision tasks using CLIP ViT-L/14, DINOv2-B/14 and for the 6 NLI tasks using Llama~3 8B.
    }
    \label{fig:radar}
\end{figure*}

\minisection{Merging Experts.}
\Cref{fig:radar} provides the per-task results of methods presented in~\cref{tab:compare_base_ours} and~\cref{tab:lora_llama_table} of the main paper. 
We observe a substantial performance improvement across tasks when merging DTE experts, compared to merging Standard-FT ones. Furthermore, when DTEs are combined using a more advanced merging strategy such as Iso-C, their per-task performance closely approaches that of the standalone experts.

\subsection{Representation Drift}
\begin{figure}[t]
    \centering
    \includegraphics[width=\columnwidth]{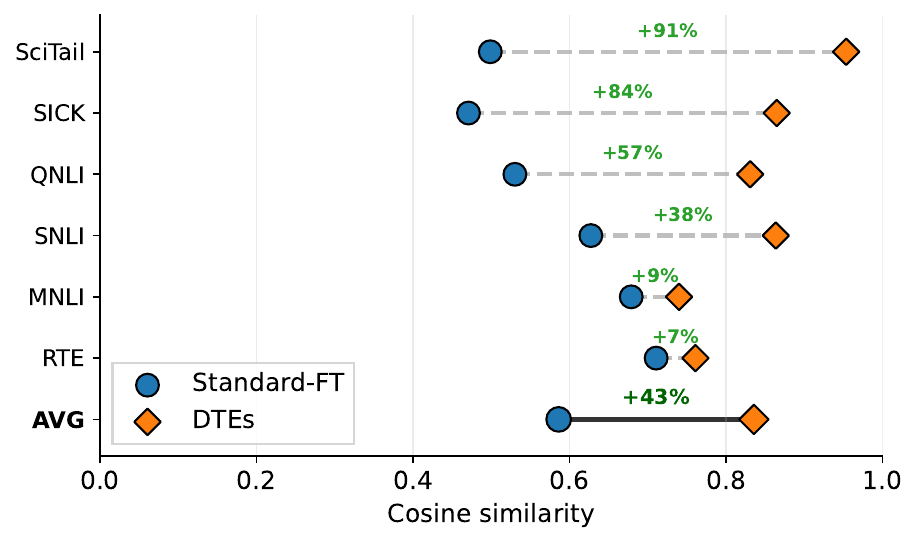}
    \caption{Cosine similarity between the pre-trained model (Llama~3 8B) and the DTEs and Standard-FT experts for the 6 NLI tasks. DTEs exhibit consistently higher similarity to the pre-trained model across all tasks, indicating smaller representation shifts after adaptation and improved compatibility for merging.}
    \label{fig:fig4llm}
\end{figure}

\minisection{Experts Drift from Pre-Trained.}
Following~\cref{sec:analysis_lbl} of the main paper, we present how much the backbone representations change during task adaptation for Llama~3 in \cref{fig:fig4llm}. As shown, DTEs consistently preserve representations closer to the pre-trained backbone. Specifically, average cosine similarity improves by approximately 43\% across the six NLI tasks, relative to Standard-FT.

\minisection{Merged Model Drift from Experts.}
We additionally measure how much the backbone representations change after merging, for both DTEs and Standard-FT experts. Specifically, we merge the experts using WA and compare the representations of the merged backbone with those of the corresponding individual experts.

Following~\cref{sec:analysis_lbl} of the main paper, let $f_{\theta_i^*}$ denote the backbone adapted to task $i$ (obtained either through Standard-FT or through the DTE procedure), and let $f_{\theta_{\text{merged}}}$ denote the merged backbone, with $\theta_{\text{merged}} = \mathcal{M}\big(\theta_0, \{\Delta\theta_i\}_{i=1}^T\big)$, where $\theta_0$ denotes the pre-trained backbone parameters, $\Delta\theta_i = \theta_i^* - \theta_0$ the task vectors, and $\mathcal{M}$ is WA. Given an input $x$, we define:
\begin{equation}
\mathbf{h}^{\text{m}}(x) = f_{\theta_{\text{merged}}}(x),
\qquad
\mathbf{h}^{(i)}(x) = f_{\theta_i^*}(x).
\end{equation}
We then measure the average cosine similarity between merged and expert representations on the test set of task $i$:
\begin{equation}
\mathrm{sim}^{\text{m}}_i =
\mathbb{E}_{x \sim \mathcal{D}_i^{\text{test}}}
\left[
\frac{f_{\theta_{\text{merged}}}(x)^\top f_{\theta_i^*}(x)}
{\|f_{\theta_{\text{merged}}}(x)\| \, \|f_{\theta_i^*}(x)\|}
\right].
\end{equation}
As in~\cref{sec:analysis_lbl} of the main paper, for CLIP we compare image embeddings.

\begin{figure}[t]
    \centering
    \includegraphics[width=0.98\columnwidth]{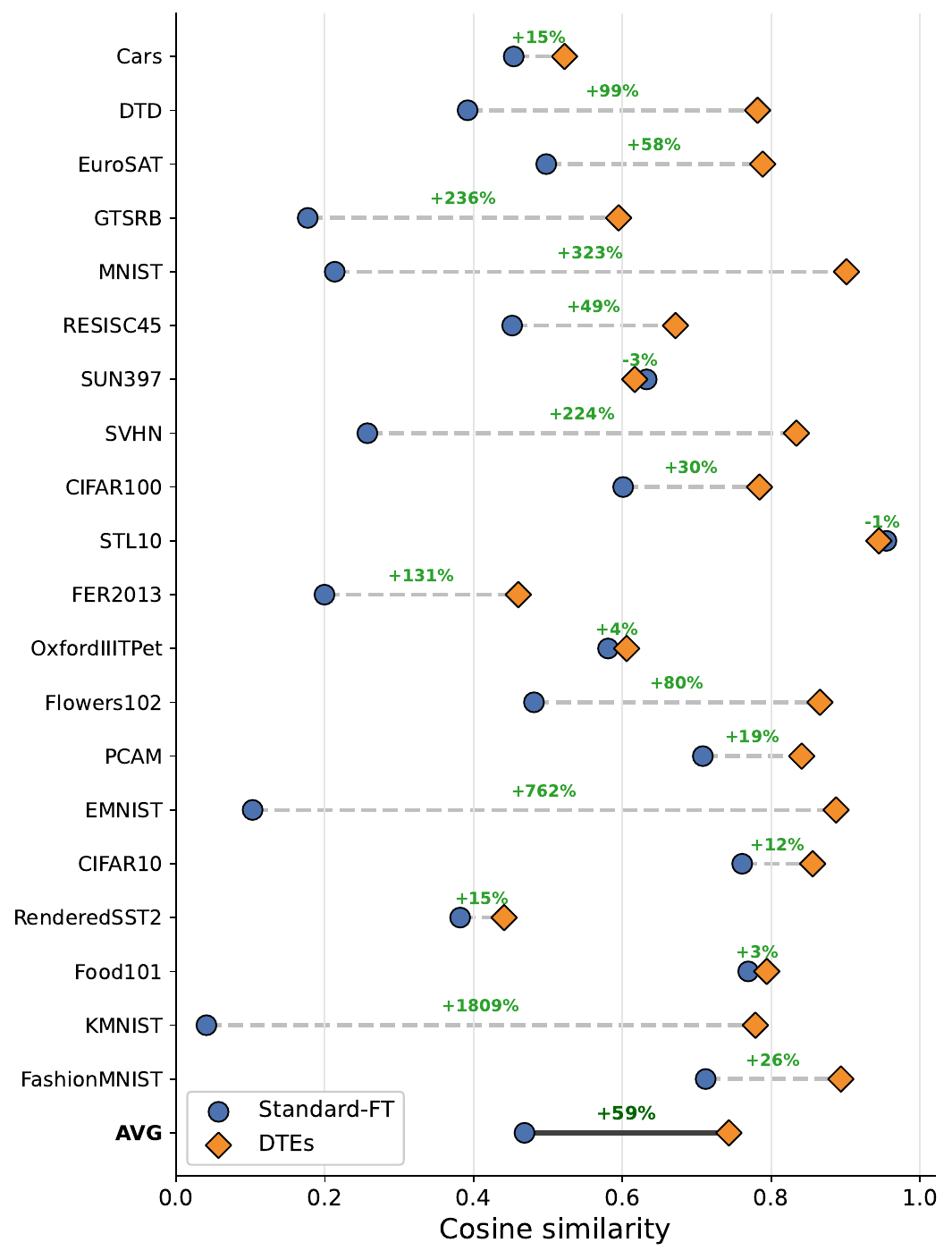}
    \caption{Cosine similarity between the individual experts and the WA-merged DTEs and Standard-FT experts for the 20-Vision benchmark and ViT-B/32. Merged DTEs exhibit consistently higher similarity to the individual experts across tasks, indicating higher representation preservation after merging, which aligns with their improved per-task performance.}
    \label{fig:reppreserve}
\end{figure}

We compute this quantity for both Standard-FT experts and DTEs.
As shown in~\cref{fig:reppreserve}, the backbone obtained by merging DTEs remains substantially closer to the individual experts, with higher cosine similarity ($\sim59\%$ relative improvement).

\subsection{DTE vs. LP-FT Training for Merging} 
In stage 2 of DTE training, we freeze the task-specific head and optimize only the backbone. In this section, we instead allow the trained task-specific head to be further co-adapted with the backbone in stage 2, similarly to LP-FT~\cite{kumar2022lpft} training. Stage 1 training, and therefore stage 2 initialization, is identical between the two methods. As displayed in~\cref{tab:lpft}, DTE training consistently demonstrates comparable or improved merged performance to LP-FT training, while also optimizing fewer parameters in stage 2 of training.

\newcommand{\nrm}[1]{$_{\scriptstyle(#1)}$}

\begin{table}[h!]
\centering
\caption{Classifier freezing (DTE) vs. co-adaptation (LP-FT) in Stage 2 of training. Average Top-1 accuracy (\%) across 6 NLI tasks using Llama~3 8B, with normalized accuracy relative to the corresponding experts in brackets.}
\label{tab:lpft}
\resizebox{\columnwidth}{!}{%
\begin{tabular}{lccc}
\toprule
\textbf{Method} & \textbf{Subspace} & \textbf{DTE} & \textbf{LP-FT} \\
\midrule
Experts         & -- & 90.89\nrm{100.0} & 90.82\nrm{100.0} \\
\midrule
Classifier-Only & -- & \multicolumn{2}{c}{78.67} \\
\midrule
WA & -- & 86.47\nrm{95.14} & 83.16\nrm{91.56} \\
\midrule
TA & -- & 89.00\nrm{97.92} & 87.86\nrm{96.75} \\
\midrule
\multirow{2}{*}{TIES}  & Full & 89.03\nrm{97.95} & 87.68\nrm{96.54} \\
\cmidrule{2-4}
                       & Core & 87.79\nrm{96.58} & 83.22\nrm{91.63} \\
\midrule
\multirow{2}{*}{TSV-M} & Full & 88.35\nrm{97.20} & 87.69\nrm{96.56} \\
\cmidrule{2-4}
                       & Core & 89.10\nrm{98.03} & 88.90\nrm{97.89} \\
\midrule
\multirow{2}{*}{Iso-C} & Full & 78.91\nrm{86.82} & 76.76\nrm{84.52} \\
\cmidrule{2-4}
                       & Core & 88.49\nrm{97.36} & 87.62\nrm{96.48} \\
\bottomrule
\end{tabular}}
\end{table}

\subsection{Performance Under Equal Computational Budget}
The computational costs for our DTE method can be slightly higher compared to Standard-FT. Therefore, in this section, we carry out an additional experiment where CLIP DTEs are trained under the same computational budget as Standard-FT CLIP experts. The implementation details for this experiment remain the same as in~\cref{sec:exp} of the main paper.

We start by calculating the total amount of backward pass FLOPs required to reach convergence for Classifier-Only adaptation (DTE Stage 1) and for Standard-FT. Results are shown in ~\cref{tab:bwd_at_best_prompt_visionft}. We observe that classifier-only adaptation requires less computations across all datasets compared to Standard-FT: up to 2 orders of magnitude less. Since Standard-FT’s FLOPs define the total training budget, we can allocate as much computation as necessary for DTE Stage 1 convergence, as it requires a strictly lower amount of FLOPs.

\begin{table}[h!]
\centering
\caption{Total number of FLOPs to convergence for Classifier-Only and Standard-FT for the 20-vision benchmark using ViT-B/32. Remaining budget corresponds to the difference between the two methods.}
\label{tab:bwd_at_best_prompt_visionft}
\resizebox{\columnwidth}{!}{
\begin{tabular}{lccc}
\toprule
 & \textbf{Classifier-Only} & \textbf{Standard-FT} & \textbf{Remaining budget} \\
\textbf{Dataset} & \textbf{(PFLOP)} & \textbf{(PFLOP)} & \textbf{(PFLOP)} \\
\midrule
Cars & 2.183 & 4.235 & 2.052 \\
DTD & 0.279 & 0.639 & 0.360 \\
EuroSAT & 0.108 & 4.405 & 4.297 \\
GTSRB & 0.518 & 4.075 & 3.557 \\
MNIST & 0.125 & 3.739 & 3.614 \\
RESISC45 & 0.504 & 4.818 & 4.314 \\
SUN397 & 4.205 & 4.251 & 0.046 \\
SVHN & 0.124 & 4.640 & 4.516 \\
CIFAR100 & 1.229 & 4.589 & 3.360 \\
STL10 & 0.061 & 4.436 & 4.375 \\
FER2013 & 0.082 & 3.513 & 3.431 \\
OxfordIIITPet & 0.297 & 2.589 & 2.292 \\
Flowers102 & 0.670 & 1.716 & 1.046 \\
PCAM & 0.024 & 4.455 & 4.431 \\
EMNIST & 0.214 & 7.987 & 7.773 \\
CIFAR10 & 0.102 & 4.589 & 4.487 \\
RenderedSST2 & 0.015 & 4.586 & 4.571 \\
Food101 & 0.975 & 4.809 & 3.834 \\
KMNIST & 0.125 & 4.673 & 4.548 \\
FashionMNIST & 0.100 & 4.673 & 4.573 \\
\midrule
\textbf{Average} & \textbf{0.597} & \textbf{4.171} & \textbf{3.574} \\
\bottomrule
\end{tabular}
}
\end{table}

As for DTE Stage 2, we use the remaining computational budget, which is defined as the FLOP difference between the two methods. This can be allocated following two different approaches:
\begin{itemize}
    \item \textbf{Approach 1 (\textit{Even budget}).}
    We allocate the same amount of computation to all tasks, defined by the average remaining budget of 3.574 PFLOPs per task.

    \item \textbf{Approach 2 (\textit{Per-dataset budget}).}
    We allocate the specific remaining budget for each task.
\end{itemize}

Depending on the approach followed, we obtain vision backbones trained using different numbers of iterations. Finally, we evaluate whether merged DTEs' capabilities are preserved under these constraints. As observed in ~\cref{tab:budget_ViT-B-32_merged}, when trained under the same computational budget as Standard-FT, DTEs achieve merging performance almost identical to that of the DTEs baseline. This verifies that the DTE methodology does not require additional training compared to typical finetuned experts.

\begin{table}[h!]
\centering
\caption{Average Top-1 accuracy (\%) across vision benchmarks using ViT-B/32, under different budget settings. DTEs (Baseline) and the corresponding merges refer to the main paper, unconstrained, DTEs. Normalized accuracy relative to the corresponding experts in brackets.}
\label{tab:budget_ViT-B-32_merged}
\setlength{\tabcolsep}{4pt}
\resizebox{\columnwidth}{!}{
\begin{tabular}{l c ccc}
\toprule
\textbf{Method} & \textbf{DTE} & \textbf{8 Tasks} & \textbf{14 Tasks} & \textbf{20 Tasks} \\
\midrule

Standard-FT 
& -- & 90.28$_{\scriptstyle(100.0)}$ & 89.31$_{\scriptstyle(100.0)}$ & 90.26$_{\scriptstyle(100.0)}$ \\

DTEs (Baseline)
& -- & 90.13$_{\scriptstyle(100.0)}$ & 89.54$_{\scriptstyle(100.0)}$ & 90.52$_{\scriptstyle(100.0)}$ \\

DTEs (Approach 1)
& -- & 89.81$_{\scriptstyle(100.0)}$ & 88.91$_{\scriptstyle(100.0)}$ & 89.81$_{\scriptstyle(100.0)}$ \\

DTEs (Approach 2)
& -- & 89.04$_{\scriptstyle(100.0)}$ & 88.73$_{\scriptstyle(100.0)}$ & 89.92$_{\scriptstyle(100.0)}$ \\

\midrule
Classifier-Only 
& -- & 78.96 & 81.01 & 81.72 \\

\midrule
WA
& \redxmark & 65.78$_{\scriptstyle(72.9)}$ & 63.51$_{\scriptstyle(71.1)}$ & 60.81$_{\scriptstyle(67.4)}$ \\

WA (Baseline)
& \greencmark & 82.47$_{\scriptstyle(91.5)}$ & 81.55$_{\scriptstyle(91.1)}$ & 81.17$_{\scriptstyle(89.7)}$ \\

WA (Approach 1)
& \greencmark & 82.43$_{\scriptstyle(91.8)}$ & 81.44$_{\scriptstyle(91.6)}$ & 81.07$_{\scriptstyle(90.3)}$ \\

WA (Approach 2)
& \greencmark & 82.41$_{\scriptstyle(92.6)}$ & 81.83$_{\scriptstyle(92.2)}$ & 81.33$_{\scriptstyle(90.4)}$ \\

\midrule
TSV-M
& \redxmark & 83.25$_{\scriptstyle(92.2)}$ & 79.60$_{\scriptstyle(89.1)}$ & 76.73$_{\scriptstyle(85.0)}$ \\

TSV-M (Baseline)
& \greencmark & 86.70$_{\scriptstyle(96.2)}$ & 84.57$_{\scriptstyle(94.5)}$ & 84.71$_{\scriptstyle(93.6)}$ \\

TSV-M (Approach 1)
& \greencmark & 86.67$_{\scriptstyle(96.5)}$ & 84.19$_{\scriptstyle(94.7)}$ & 84.39$_{\scriptstyle(94.0)}$ \\

TSV-M (Approach 2)
& \greencmark & 85.85$_{\scriptstyle(96.4)}$ & 84.23$_{\scriptstyle(94.9)}$ & 84.26$_{\scriptstyle(93.7)}$ \\

\bottomrule
\end{tabular}
}
\end{table}

\subsection{Joint-Task Evaluation: Classifier Construction}
We provide further details on the construction of the joint classification head for the 8-Vision benchmark, used in the joint-task evaluation of~\cref{sec:mergevision} of the main paper. 

Following~\cite{stoica2024knots}, the head is defined over the union of class labels of all eight tasks, with duplicate labels removed.
For Standard-FT experts, class prototypes are obtained from handcrafted prompts. Since MNIST and SVHN share identical class names and prompt templates, their class prototypes coincide and each digit class appears only once.
For DTEs, each class prototype is obtained with the learned prompt of its corresponding task. As the learned prompts of MNIST and SVHN differ, the two datasets yield different prototypes for the same digit class. To keep the label space identical across the two settings, we construct a single prototype for each shared digit class by averaging the corresponding MNIST and SVHN prototypes. In both settings, the joint head therefore contains the same 748 target classes.

\subsection{Prompt Context Tokens}
In the main paper, we use $M=16$ context tokens for all CLIP experiments, following the default configuration of CoOp.

\begin{figure*}[h!]
    \centering
    \includegraphics[width=0.84\textwidth]{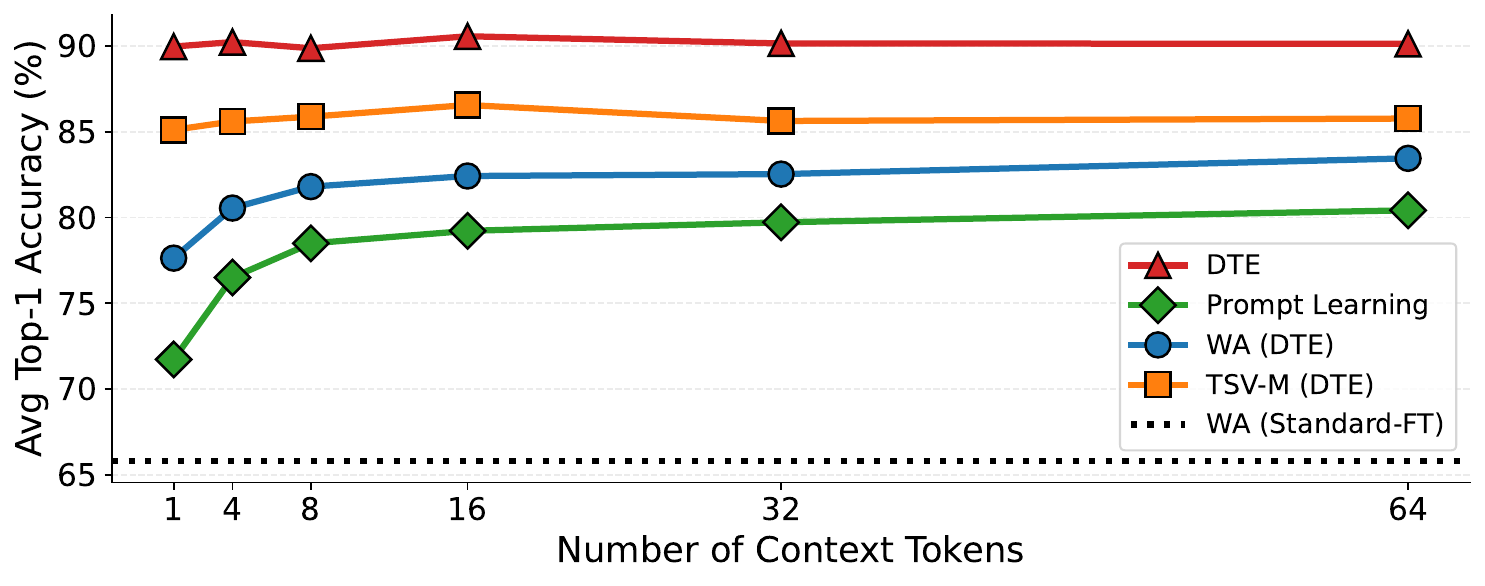}
    \caption{Average Accuracy (\%) on the 8-Vision benchmark as a function of the number of context tokens for ViT-B/32 across different methods.
    }
    \label{fig:ctx8vis}
\end{figure*}

In this section we perform an ablation study on the number of context tokens used during prompt learning. \Cref{fig:ctx8vis} reports results on the 8-Vision benchmark, for $M \in \{1, 4, 8, 16, 32, 64\}$, across different methods.

We observe that Prompt Learning with even a single context token already outperforms the WA (Standard-FT) merging by a notable margin. As the number of context tokens increases, the performance of both Prompt Learning and DTE Merging improves, eventually saturating around $M=16$ tokens.

\subsection{Prompt Learning Memory Overhead} 
\Cref{tab:prompt_overhead} compares the number of parameters of the CLIP encoders with the additional parameters introduced by Prompt Learning for a single task. The results show that Prompt Learning adds only a negligible number of parameters compared to the full model, resulting in minimal memory overhead. The number of per-task prompt learning parameters is calculated as: \[
\text{Prompt Params} = M \times d
\]
where $M$ denotes the number of context tokens and $d$ is the dimensionality of the CLIP word embeddings (512 for ViT-B/32 and 768 for ViT-L/14).

\begin{table}[h!]
\centering
\caption{Parameters comparison between CLIP model and Prompt Learning with $M=16$ context tokens. Percentage (\%) is computed as $(\text{Prompt Params} / \text{Total Params}) \times 100$.}
\label{tab:prompt_overhead}
\setlength{\tabcolsep}{5pt}
\resizebox{\columnwidth}{!}{
\begin{tabular}{lccc}
\toprule
\textbf{Model} & \textbf{Total Params} & \textbf{Prompt Params} &\textbf{Percentage}\\
\midrule
ViT-B/32 & 151 M & 8 K & 0.005 \%    \\
ViT-L/14 & 428 M & 12 K & 0.003 \% \\
\bottomrule
\end{tabular}
}
\end{table}

\subsection{DTE Sensitivity to Prompt Initialization}
Learnable prompts trained during CLIP DTE Stage 1 are randomly initialized following the default configuration of CoOp~\cite{zhou2022learning}. This adds more stochasticity to the training pipeline, which could potentially lead to high variance in performance across experiments. However, we find that this is not the case. We conduct 3 full runs of our method for ViT-B/32 using different seeds and report their mean accuracy together with the sample standard deviation in~\cref{tab:3runs}. We observe that standard deviations are systematically below 0.5 across all reported settings, suggesting that prompt initializations introduce limited additional stochasticity, therefore showcasing the consistency of our method. 

\begin{table}[h!]
\centering
\caption{Average Top-1 accuracy (\%) across vision benchmarks for ViT-B/32. Results are reported as mean $\pm$ standard deviation over 3 full runs of our method.}
\label{tab:3runs}
\setlength{\tabcolsep}{3pt}
\resizebox{\columnwidth}{!}{%
\begin{tabular}{l c ccc}
\toprule
& & \multicolumn{3}{c}{\textbf{ViT-B/32}} \\
\cmidrule(lr){3-5}
\textbf{Method} & \textbf{DTE} & \textbf{8 Tasks} & \textbf{14 Tasks} & \textbf{20 Tasks} \\
\midrule
Zero-shot & -- & 48.24 & 56.43 & 55.56 \\
\midrule
DTEs & -- & 90.21$_{\scriptstyle\pm 0.33}$ & 89.58$_{\scriptstyle\pm 0.23}$ & 90.49$_{\scriptstyle\pm 0.21}$ \\
\midrule
Prompt learning & -- & 79.01$_{\scriptstyle\pm 0.18}$ & 81.05$_{\scriptstyle\pm 0.13}$ & 81.66$_{\scriptstyle\pm 0.07}$ \\
\midrule
WA & \greencmark & 82.49$_{\scriptstyle\pm 0.06}$ & 81.51$_{\scriptstyle\pm 0.18}$ & 81.02$_{\scriptstyle\pm 0.25}$ \\
\midrule
TA & \greencmark & 82.50$_{\scriptstyle\pm 0.07}$ & 81.88$_{\scriptstyle\pm 0.07}$ & 81.02$_{\scriptstyle\pm 0.25}$ \\
\midrule
TIES & \greencmark & 83.30$_{\scriptstyle\pm 0.23}$ & 81.68$_{\scriptstyle\pm 0.39}$ & 82.22$_{\scriptstyle\pm 0.10}$ \\
\midrule
TSV-M & \greencmark & 86.86$_{\scriptstyle\pm 0.44}$ & 84.48$_{\scriptstyle\pm 0.11}$ & 84.47$_{\scriptstyle\pm 0.33}$ \\
\midrule
Iso-C & \greencmark & 87.05$_{\scriptstyle\pm 0.46}$ & 84.45$_{\scriptstyle\pm 0.24}$ & 84.18$_{\scriptstyle\pm 0.19}$ \\
\bottomrule
\end{tabular}%
}
\end{table}

\subsection{CLIP Classifier Adaptation via Text Encoder Fine-Tuning}
Throughout the paper, we use prompt learning as a parameter-efficient way to adapt the text encoder-based CLIP classifier in Stage~1 of DTE training. Here, we instead fine-tune the full text encoder in Stage~1. In Stage~2, as before, the trained per-task classifiers are kept frozen and only the vision backbone is trained.

\Cref{tab:textenc} reports the merging results on the 8-Vision benchmark. As expected, full text encoder fine-tuning in Stage~1 yields stronger classifiers (82.80\% vs.\ 78.96\% classifier-only accuracy) and moderately improves merged accuracy over prompt-learned DTEs.

\begin{table}[h!]
\centering
\caption{DTEs with prompt learning vs.\ full text encoder fine-tuning in Stage 1. Average Top-1 accuracy (\%) on the 8-Vision benchmark using ViT-B/32, with normalized accuracy relative to the corresponding experts in brackets.}
\label{tab:textenc}
\resizebox{\columnwidth}{!}{%
\begin{tabular}{lcc}
\toprule
\textbf{Method} & \textbf{DTE} & \textbf{DTE (TE-FT)} \\
\midrule
Experts         & 90.13\nrm{100.0} & 90.37\nrm{100.0} \\
\midrule
Classifier-Only & 78.96 & 82.80 \\
\midrule
WA    & 82.47\nrm{91.5} & 84.86\nrm{93.9} \\
\midrule
TA    & 82.47\nrm{91.5} & 84.86\nrm{93.9} \\
\midrule
TIES  & 83.05\nrm{92.1} & 85.34\nrm{94.4} \\
\midrule
TSV-M & 86.70\nrm{96.2} & 87.80\nrm{97.2} \\
\midrule
Iso-C & 86.59\nrm{96.1} & 87.45\nrm{96.8} \\
\bottomrule
\end{tabular}}
\end{table}
However, fully training the text encoder is more expensive, and requires the storing and distribution of either a full text encoder per task or a fixed set of precomputed class prototypes. We therefore use prompt learning as the default, and view text encoder fine-tuning as a viable alternative when compute is available and task-specific class prototypes can be shared directly.

\section{Handcrafted Templates}
\label{sec:supp_templates}

Throughout the paper, the zero-shot text encoder-based classification head of a CLIP model is built from an ensemble of handcrafted templates, following previous model merging literature~\cite{ilharco2023editing,tsv}. 

The exact template sets of all $20$ vision datasets are listed in~\cref{tab:supp_templates}.

\begin{table*}[t]
\centering
\caption{
Handcrafted templates used to build the zero-shot text encoder-based CLIP classification head for each dataset.
The placeholder \texttt{\{c\}} is replaced by the classname,
``\#'' denotes the number of templates in the ensemble,
and templates within a row are separated by ``$\vert$''.
Rows listing several datasets indicate that those datasets
share an identical template set.
}
\label{tab:supp_templates}
\resizebox{0.68\textwidth}{!}{%
\begin{tabular}{
    @{}
    l
    c
    >{\ttfamily\raggedright\arraybackslash}p{0.82\textwidth}
    @{}
}
\toprule
\textbf{Dataset}
&
\textbf{\#}
&
\textnormal{\textbf{Templates}}
\\
\midrule

Cars & 8 &
a photo of a \{c\}. $\vert$
a photo of the \{c\}. $\vert$
a photo of my \{c\}. $\vert$
i love my \{c\}! $\vert$
a photo of my dirty \{c\}. $\vert$
a photo of my clean \{c\}. $\vert$
a photo of my new \{c\}. $\vert$
a photo of my old \{c\}.
\\

\addlinespace[2pt]

DTD & 8 &
a photo of a \{c\} texture. $\vert$
a photo of a \{c\} pattern. $\vert$
a photo of a \{c\} thing. $\vert$
a photo of a \{c\} object. $\vert$
a photo of the \{c\} texture. $\vert$
a photo of the \{c\} pattern. $\vert$
a photo of the \{c\} thing. $\vert$
a photo of the \{c\} object.
\\

\addlinespace[2pt]

EuroSAT & 3 &
a centered satellite photo of \{c\}. $\vert$
a centered satellite photo of a \{c\}. $\vert$
a centered satellite photo of the \{c\}.
\\

\addlinespace[2pt]

GTSRB & 3 &
a zoomed in photo of a "\{c\}" traffic sign. $\vert$
a centered photo of a "\{c\}" traffic sign. $\vert$
a close up photo of a "\{c\}" traffic sign.
\\

\addlinespace[2pt]

MNIST, SVHN & 1 &
a photo of the number: "\{c\}".
\\

\addlinespace[2pt]

RESISC45 & 18 &
satellite imagery of \{c\}. $\vert$
aerial imagery of \{c\}. $\vert$
satellite photo of \{c\}. $\vert$
aerial photo of \{c\}. $\vert$
satellite view of \{c\}. $\vert$
aerial view of \{c\}. $\vert$
satellite imagery of a \{c\}. $\vert$
aerial imagery of a \{c\}. $\vert$
satellite photo of a \{c\}. $\vert$
aerial photo of a \{c\}. $\vert$
satellite view of a \{c\}. $\vert$
aerial view of a \{c\}. $\vert$
satellite imagery of the \{c\}. $\vert$
aerial imagery of the \{c\}. $\vert$
satellite photo of the \{c\}. $\vert$
aerial photo of the \{c\}. $\vert$
satellite view of the \{c\}. $\vert$
aerial view of the \{c\}.
\\

\addlinespace[2pt]

SUN397, STL10 & 2 &
a photo of a \{c\}. $\vert$
a photo of the \{c\}.
\\

\addlinespace[2pt]

CIFAR100, CIFAR10 & 18 &
a photo of a \{c\}. $\vert$
a blurry photo of a \{c\}. $\vert$
a black and white photo of a \{c\}. $\vert$
a low contrast photo of a \{c\}. $\vert$
a high contrast photo of a \{c\}. $\vert$
a bad photo of a \{c\}. $\vert$
a good photo of a \{c\}. $\vert$
a photo of a small \{c\}. $\vert$
a photo of a big \{c\}. $\vert$
a photo of the \{c\}. $\vert$
a blurry photo of the \{c\}. $\vert$
a black and white photo of the \{c\}. $\vert$
a low contrast photo of the \{c\}. $\vert$
a high contrast photo of the \{c\}. $\vert$
a bad photo of the \{c\}. $\vert$
a good photo of the \{c\}. $\vert$
a photo of the small \{c\}. $\vert$
a photo of the big \{c\}.
\\

\addlinespace[2pt]

FER2013 & 6 &
a photo of a \{c\} looking face. $\vert$
a photo of a face showing the emotion: \{c\}. $\vert$
a photo of a face looking \{c\}. $\vert$
a face that looks \{c\}. $\vert$
they look \{c\}. $\vert$
look at how \{c\} they are.
\\

\addlinespace[2pt]

OxfordIIITPet & 1 &
a photo of a \{c\}, a type of pet.
\\

\addlinespace[2pt]

Flowers102 & 1 &
a photo of a \{c\}, a type of flower.
\\

\addlinespace[2pt]

PCAM & 1 &
this is a photo of \{c\}
\\

\addlinespace[2pt]

EMNIST & 1 &
a photo of the digit character: "\{c\}".
\\

\addlinespace[2pt]

RenderedSST2 & 1 &
a \{c\} review of a movie.
\\

\addlinespace[2pt]

Food101 & 1 &
a photo of \{c\}, a type of food.
\\

\addlinespace[2pt]

KMNIST & 1 &
a photo of the character \{c\}.
\\

\addlinespace[2pt]

FashionMNIST & 2 &
a photo of a \{c\}. $\vert$
a photo of the \{c\}.
\\

\bottomrule
\end{tabular}%
}
\end{table*}

\end{document}